\documentclass[final,1p,times]{elsarticle}

\usepackage{mathptmx} % assumes new font selection scheme installed
\usepackage{times} % assumes new font selection scheme installed
\usepackage{amsmath} % assumes amsmath package installed
\usepackage{amssymb}  % assumes amsmath package installed
\usepackage{float} % to allow [H] float specifier
\usepackage{subcaption}
\usepackage{tabularx}
\usepackage{float}
\usepackage{titlesec}
\usepackage{tabularx}
\usepackage{placeins}
\usepackage[labelsep=period]{caption}
\usepackage{multirow}
\usepackage{indentfirst}
\usepackage{xpatch}
\usepackage{setspace}
\usepackage{mathptmx} 
\usepackage{booktabs}
\usepackage{url}

\usepackage{subcaption}
\usepackage{epsfig}
\usepackage{tikz}
\journal{Aerospace Science and Technology}

\begin{document}

\begin{frontmatter}



\title{QP-Based Control of an Underactuated Aerial Manipulator under Constraints} %% Article title


 \author[label1,label2]{Nesserine LARIBI}
 \author[label1,label2]{Mohammed Rida Mokhtari}
 \author[label3]{Abdelaziz Benallegue}
 \author[label3]{Abdelhafid El-Hadri}
 \author[label4]{Mehdi Benallegue}
 \affiliation[label1]{organization={Ecole superieure en sciences appliquees de tlemcen ESSAT},             
          city={Tlemcen},
            postcode={13000},
            country={Algeria}}

\affiliation[label2]{organization={Laboratoire d'automatique de Tlemcen},
         city={Tlemcen},
            postcode={13000},
            country={Algeria}}
\affiliation[label3]{organization={Laboratoire d’Ingénierie des Systèmes de Versailles, Université de Versailles Saint Quentin en Yvelines, France},
         city={Versailles},
            postcode={78000},
            country={France}}
\affiliation[label4]{organization={ Joint Robotics Laboratory, National Institute of Advanced Industrial Science and Technology (AIST)},
         city={Tsukuba},
            country={Japan}}
\begin{abstract}
This paper presents a constraint-aware control framework for underactuated aerial manipulators, enabling accurate end-effector trajectory tracking while explicitly accounting for safety and feasibility constraints. The control problem is formulated as a quadratic program that computes dynamically consistent generalized accelerations subject to underactuation, actuator bounds, and system constraints. To enhance robustness against disturbances, modeling uncertainties, and steady-state errors, a passivity-based integral action is incorporated at the torque level without compromising feasibility. The effectiveness of the proposed approach is demonstrated through high-fidelity physics-based simulations, including parameter perturbations, viscous joint friction, and realistic sensing and state-estimation effects, showing accurate tracking, smooth control inputs, and reliable constraint satisfaction under realistic operating conditions.
\end{abstract}

\begin{graphicalabstract}
\end{graphicalabstract}

\begin{highlights}
 \item Constraint-aware dynamic modeling of an underactuated aerial manipulator with a floating base and onboard manipulator.
  
  \item Whole-body control formulation enabling coordinated regulation of the aerial platform and the manipulator.
  
  \item Quadratic programming (QP)–based control framework explicitly handling underactuation, actuator limits, and task feasibility.
  
  \item Geometric thrust-direction regulation integrated with the whole-body QP to address quadrotor underactuation.
  
  \item Passivity-based integral action for enhanced disturbance rejection and steady-state accuracy.
  
  \item Unified control of base position, yaw orientation, and end-effector motion in task space.
  
  \item Robust set-point stabilization and trajectory tracking under significant model uncertainties.
  
  \item High-fidelity physics-based simulations demonstrating real-time feasibility and improved tracking performance.

\end{highlights}

%% Keywords

\end{frontmatter}
%%Aerial manipulators, which integrate an aerial platform with a robotic arm or gripper, have emerged as a promising class of robotic systems that extend the role of aerial robots from passive observation to active physical interaction with the environment \cite{suarez2018design, tognon2019truly}. By combining the mobility and three-dimensional maneuverability of aerial vehicles with the dexterity of robotic manipulators, these systems enable contact-rich tasks such as grasping, manipulation, and assembly in locations that are otherwise difficult or dangerous for humans to access. The aerial platform provides rapid mobility and precise positioning in space, while the onboard manipulator allows the system to physically interact with its surroundings. This synergy unlocks new mission capabilities for aerial robots, transforming them from flying sensors into versatile robotic workers capable of intervention.
\section{Introduction}
Aerial manipulators (AMs) have emerged as a significant advancement in aerial robotics, extending the role of unmanned aerial systems from passive observation to active physical interaction with the environment \cite{suarez2018design, tognon2019truly}. Unlike terrestrial robotic platforms, AMs are not constrained by terrain accessibility, and unlike conventional unmanned aerial vehicles (UAVs), they are no longer limited to remote sensing. Instead, they enable true three‑dimensional physical interaction, greatly expanding the range of tasks that can be performed in previously inaccessible or hazardous environments.

The potential applications of aerial manipulators continue to grow with technological progress. In industrial inspection and maintenance, they can carry specialized tools for non‑destructive testing or servicing of power lines, bridges, and large infrastructures \cite{korpela2014towards, bodie2020active}. In agriculture and environmental monitoring, they have been deployed for fruit harvesting, sample collection, and canopy inspection, where hovering and physical interaction are essential \cite{suarez2018design}. Additional envisioned applications include search and rescue, disaster response, and hazardous material handling, where AMs can operate in confined or unsafe areas without exposing human operators to risk.

Aerial manipulators can be categorized according to the actuation characteristics of the aerial platform and the kinematic structure of the attached manipulator. Platforms are typically either underactuated, such as multirotors~\cite{ruggiero2015multilayer} and helicopters~\cite{pounds2014stability}, which exhibit inherently coupled position–attitude dynamics, or fully actuated, such as tilted‑rotor hexacopters~\cite{bodie2020active} and omnidirectional platforms~\cite{kamel2018voliro}, which allow independent control of translation and orientation~\cite{ruggiero2018aerial}. Manipulators vary in degrees of freedom, joint types, and kinematic configuration~\cite{wei2021review}, ranging from lightweight arms to redundant mechanisms, depending on task requirements and payload constraints.
A key challenge in aerial manipulation lies in the strong dynamic coupling between the UAV and the manipulator, particularly during precise end‑effector motion or physical interaction. Existing control strategies for underactuated aerial manipulators can be broadly classified into decoupled and coupled approaches. Decoupled methods treat the UAV and the manipulator as separate subsystems with independently designed controllers~\cite{bulut2019decoupled, zhang2019robust, lippiello2012exploiting}. While computationally efficient, these approaches often neglect internal coupling effects, leading to degraded performance in dynamic or contact‑rich tasks. Coupled approaches instead model and control the aerial manipulator as a unified dynamical system~\cite{jeong2023passivity, samadikhoshkho2020nonlinear, lippiello2012cartesian}, enabling coordinated behavior and improved accuracy during aggressive maneuvers or physical interaction when nonlinear dynamics are accurately modeled. However, most existing coupled controllers do not explicitly enforce hard physical constraints, such as actuator saturation, input bounds, or state limits, despite their importance for real‑world feasibility.

Enforcing such constraints is crucial for guaranteeing stability and safe operation in practice~\cite{nava2019direct}. Quadratic‑programming (QP)–based schemes have been explored for contact‑force regulation and multi‑task optimization~\cite{nava2020direct}, yet these formulations typically assume a fully actuated base and do not incorporate the underactuated nature of multirotor platforms.

This work addresses this gap by proposing a centralized, constraint‑aware control framework for underactuated aerial manipulators. The controller regulates the floating‑base position, yaw orientation, and end‑effector motion relative to the base, while explicitly accounting for platform underactuation and enforcing actuator limits for both the aerial base and the manipulator. A passivity‑based integral action is incorporated at the torque‑control level to enhance robustness against modeling uncertainties and sensor noise. To the best of our knowledge, such a formulation has not been previously investigated for underactuated aerial manipulators, although related QP‑based schemes with integral terms have been explored in humanoid robotics~\cite{cisneros2018robust}.

The main contributions of this work are summarized as follows:
\begin{itemize}
    \item A whole-body QP‑based controller that computes dynamically consistent generalized accelerations, enabling feasible tasks execution while accounting for the full coupled dynamics of the aerial manipulator.
    \item A passivity‑based integral term introduced at the torque‑control level to improve robustness to modeling errors and measurement noise.
\end{itemize}

A high‑fidelity, physics‑based simulation is used 
to validate the proposed controller demonstrating 
significant improvements in end‑effector tracking 
accuracy. The simulation environment includes model 
uncertainties, joint friction, and sensor noise to 
closely approximate real‑world operating conditions 
and assess the robustness of the approach.

The remainder of this paper is organized as 
follows. Section II presents the system modeling. 
Section III details the proposed control framework. 
Section IV reports the simulation setup and results. 
Section V concludes the paper and outlines directions for future research.
\section{Modeling of the Aerial Manipulator}
\begin{figure}[H]
    \centering
    \includegraphics[width=0.6\columnwidth]{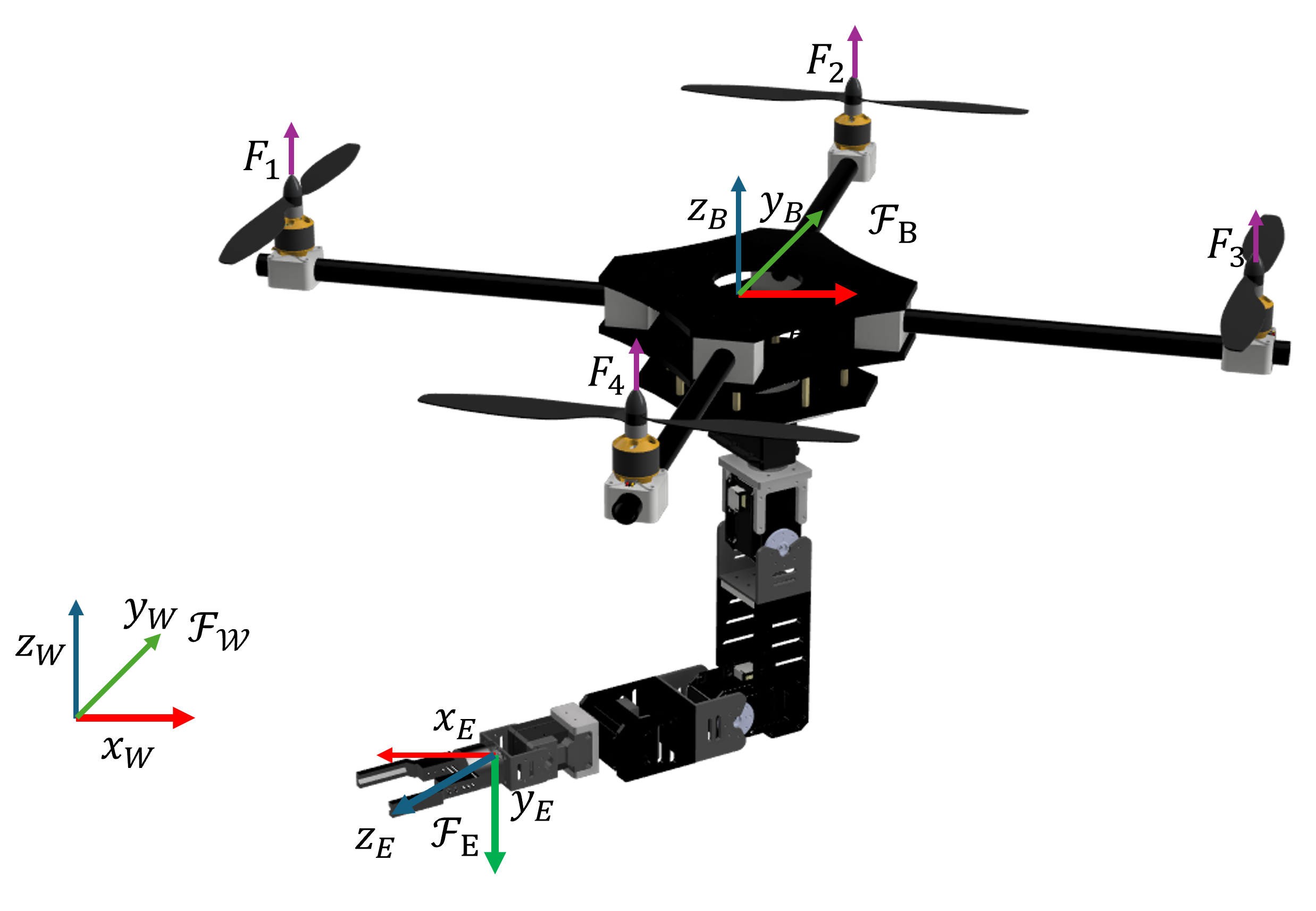}
    \caption{CAD model of the aerial manipulator consisting of a quadrotor base and a 5--DoF arm.}
    \label{fig:system_frames}
\end{figure}

The aerial manipulator consists of a quadrotor UAV acting as a floating base, rigidly equipped with a serial $n$--DoF robotic arm, as illustrated in Fig.~\ref{fig:system_frames}. Three reference frames are defined: the inertial frame $\mathcal{F}_W$, the body frame $\mathcal{F}_B$ attached to the quadrotor center of mass, and the end--effector frame $\mathcal{F}_E$ fixed to the terminal link of the manipulator.

\subsection{Dynamic Modeling}

The configuration of the aerial manipulator is described by the position
$p_B \in \mathbb{R}^3$ and orientation $R \in SO(3)$ of the quadrotor base with respect to $\mathcal{F}_W$, together with the joint coordinates of the manipulator
$\mathbf{q}_M \in \mathbb{R}^n$. The complete configuration space is thus given by
\begin{equation}
\mathbf{q} = (p_B, R, \mathbf{q}_M)
\;\in\;
\mathbb{R}^3 \times SO(3) \times \mathbb{R}^n .
\end{equation}

The corresponding generalized velocity vector is defined as
\begin{equation}
\boldsymbol{\nu} =
\begin{bmatrix}
\dot{p}_B^\top &
\omega_B^\top &
\dot{\mathbf{q}}_M^\top
\end{bmatrix}^\top
\in \mathbb{R}^{6+n},
\end{equation}
The position of the center of mass (CoM) and the orientation of the body-attached reference frame of the $i$--th rigid body, expressed in the inertial frame $\mathcal{F}_W$, are given by
\begin{align}
p_{i,c} &= p_B + R\,p_{i,c}^B, \\
R_{i} &= R\,R_{i}^B .
\end{align}
Here, $p_{i,c}^B$ and $R_{i}^B$ denote the CoM position and the orientation of the body-attached frame of the $i$--th rigid body, expressed with respect to the base frame $\mathcal{F}_B$.

The corresponding linear and angular velocities, expressed in the inertial frame $\mathcal{F}_W$ and the body-attached frame $\mathcal{F}_{i,c}$, respectively, follow from the kinematic relation
\begin{equation}
\begin{bmatrix}
\dot{p}_{i,c} \\
\omega_{i,c}
\end{bmatrix}
=
\begin{bmatrix}
\mathbf{I}_3 & -R\,S(p_{i,c}^B) & R\,J_{v_{i,c}}^B \\
\mathbf{0}_{3} & R_{i}^{B\top} & R_{i}^{B\top} J_{\omega_i}^B
\end{bmatrix}
\boldsymbol{\nu},
\end{equation}
where $S(\cdot)$ denotes the skew-symmetric operator, and
$J_{v_{i,c}}^B$ and $J_{\omega_i}^B$ are the translational and rotational Jacobians associated with the CoM, expressed in the base frame $\mathcal{F}_B$.

For compactness, this relation can be rewritten as
\begin{equation}
\begin{bmatrix}
\dot{p}_{i,c} \\
\omega_{i,c}
\end{bmatrix}
=
\begin{bmatrix}
J_{v_{i,c}} \\
J_{\omega_i}
\end{bmatrix}
\boldsymbol{\nu},
\end{equation}
where $J_{v_{i,c}}$ and $J_{\omega_i}$ denote the corresponding translational and rotational Jacobians.

\subsection{Equations of Motion}

The aerial manipulator exhibits strong dynamic coupling between the serial manipulator and the underactuated aerial platform. While classical Euler–Lagrange or Newton–Euler formulations can be applied~\cite{kannan2013modeling}, the explicit derivation of Coriolis and centrifugal terms quickly becomes analytically cumbersome as the system complexity increases. 

To obtain a rigorous yet tractable representation, the equations of motion are derived using Gauss’s principle under standard rigid-body assumptions, while neglecting aerodynamic effects acting on the manipulator~\cite{wieber2006holonomy}. The resulting dynamics can be written in the compact form
\begin{equation}
    M(\mathbf{q})\,\dot{\boldsymbol{\nu}}
    +
    C(\mathbf{q},\boldsymbol{\nu})\,\boldsymbol{\nu}
    +
    G(\mathbf{q})
    =
    \mathbf{u},
    \label{eq:gauss_dynamics}
\end{equation}
where $M(\mathbf{q}) \in \mathbb{R}^{(6+n)\times(6+n)}$ is the symmetric positive-definite mass matrix,
$C(\mathbf{q},\boldsymbol{\nu}) \in \mathbb{R}^{(6+n)\times(6+n)}$ collects Coriolis and centrifugal effects and satisfies the standard skew-symmetry property $\dot{M} - 2C$, and
$G(\mathbf{q}) \in \mathbb{R}^{6+n}$ is the gravitational vector.

The matrices $M(\mathbf{q})$, $C(\mathbf{q},\boldsymbol{\nu})$, and $G(\mathbf{q})$ are assembled by summing the contributions of all rigid bodies composing the system:
\begin{equation}
\begin{cases}
\displaystyle
M = \sum_{i}
\left(
J_{v_{i,c}}^{\top} m_i J_{v_{i,c}}
+
J_{\omega_i}^{\top} \mathbb{I}_i J_{\omega_i}
\right), \\[8pt]
\displaystyle
C = \sum_{i}
\left(
J_{v_{i,c}}^{\top} m_i \dot{J}_{v_{i,c}}
+
J_{\omega_i}^{\top} \mathbb{I}_i \dot{J}_{\omega_i}
-
J_{\omega_i}^{\top} S\!\left(\mathbb{I}_i J_{\omega_i}\boldsymbol{\nu}\right) J_{\omega_i}
\right), \\[10pt]
\displaystyle
G = \sum_{i} J_{v_{i,c}}^{\top} f_i ,
\end{cases}
\label{eq:composite_terms}
\end{equation}
where $m_i$ and $\mathbb{I}_i$ denote the mass and inertia tensor of the $i$-th rigid body, respectively. The gravitational force acting on body $i$ is given by
\begin{equation}
    f_i = \begin{bmatrix} 0 \\ 0 \\ -m_i g \end{bmatrix}.
\end{equation}

The generalized input vector is defined as
\begin{equation}
\mathbf{u} =
\begin{bmatrix}
\mathbf{w}_B^\top &
\boldsymbol{\tau}_M^\top
\end{bmatrix}^\top
\in \mathbb{R}^{6+n},
\end{equation}
where $\mathbf{w}_B \in \mathbb{R}^{6}$ is the wrench applied to the floating base and $\boldsymbol{\tau}_M \in \mathbb{R}^{n}$ are the joint torques generated by the manipulator actuators.

\subsubsection{Quadrotor Wrench Mapping}

For the quadrotor base, the applied wrench is generated by the total 
thrust $f_{z,b} \in \mathbb{R}^{+}$ and the body 
torques $\boldsymbol{\tau}_b \in \mathbb{R}^{3}$, 
both expressed in the body frame $\mathcal{F}_B$. Their mapping to the generalized base wrench is written as
\begin{equation}
    \mathbf{w}_B
    =
    \underbrace{
    \begin{bmatrix}
        R e_3 & \mathbf{0}_{3\times 3} \\
        \mathbf{0}_{3\times 1} & \mathbf{I}_3
    \end{bmatrix}}_{T_w}
    \begin{bmatrix}
        f_{z,b} \\[3pt]
        \boldsymbol{\tau}_b
    \end{bmatrix}
    =
    T_w\,\Xi\,\mathbf{F},
    \label{eq:wrench_allocation}
\end{equation}
where $e_3 = [0\;0\;1]^\top$,
$\mathbf{F} = [F_1\;F_2\;F_3\;F_4]^\top$ collects the individual rotor thrusts ($F_i \ge 0$), and
$\Xi \in \mathbb{R}^{4\times4}$ is the allocation matrix mapping rotor thrusts to the collective thrust and body torques.

\section{Control Design}

\begin{figure}[h]
  \centering
  \includegraphics[width=0.85\linewidth]{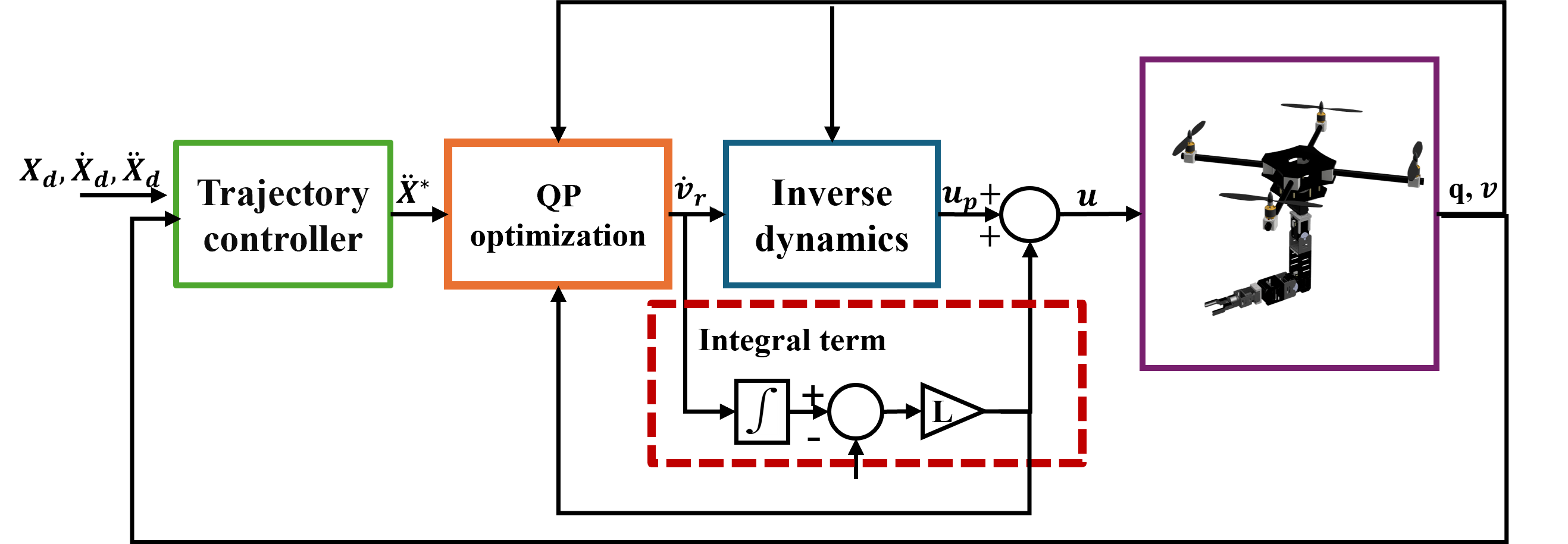}
  \caption{Block diagram of the proposed control architecture.}
  \label{fig:control_structure}
\end{figure}

The control objective is to track the Cartesian position and yaw orientation of the floating base, together with the end--effector position expressed in the base frame $\mathcal{F}_B$. As illustrated in Fig.~\ref{fig:control_structure}, the proposed controller adopts a two--loop hierarchical architecture. The outer loop operates in task space and generates a desired task--space acceleration $\ddot{X}^*$ based on tracking errors, while the inner loop enforces dynamic consistency by computing generalized accelerations and control inputs that best realize $\ddot{X}^*$ subject to the system dynamics and actuation constraints. To improve robustness against modeling uncertainties and unmodeled disturbances, an integral action is introduced at the torque level.

\subsection{Outer-Loop Trajectory Controller}

We define the desired task trajectory as:
\begin{equation}
X_{d} \;=\; \big(\,\mathbf{p}_{B,d},\, \mathbf{X}_{\psi,d},\, \mathbf{p}_{E,d}^{B}\,\big),
\qquad
\mathbf{X}_{\psi,d} \;=\; R_d^\top e_1,
\end{equation}
The desired quantities $\mathbf{p}_{B,d} \in \mathbb{R}^3$, $\mathbf{p}_{E,d}^{B} \in \mathbb{R}^3$, and $\mathbf{X}_{\psi,d} \in \mathbb{R}^3$ represent the base 
position in $\mathcal{F}_W$, 
the end-effector position in $\mathcal{F}_B$, and a unit vector encoding the desired yaw orientation, respectively, defined as
\begin{equation}
\mathbf{X}_{\psi} \;=\; R^\top e_1,\quad e_1 = [1\;0\;0]^\top
\end{equation}

where $R_d$ denotes the desired base orientation.

A second--order task--space tracking law is employed to generate the desired task acceleration $\ddot{X}^* \in \mathbb{R}^9$:
\begin{equation}
\ddot{X}^* \;=\;
\begin{bmatrix}
\ddot{\mathbf{p}}_{B}^{*} \\[4pt]
\ddot{\mathbf{X}}_{\psi}^{*} \\[4pt]
\ddot{\mathbf{p}}_{E}^{B\,*}
\end{bmatrix}
=
\begin{bmatrix}
\ddot{\mathbf{p}}_{B,d}
+ K_{v,B} (\dot{\mathbf{p}}_{B,d}-\dot{\mathbf{p}}_{B})
+ K_{p,B} (\mathbf{p}_{B,d}-\mathbf{p}_{B})
\\[6pt]
\ddot{\mathbf{X}}_{\psi,d}
+ K_{v,\psi} (\dot{\mathbf{X}}_{\psi,d}-\dot{\mathbf{X}}_{\psi})
+ K_{p,\psi} (\mathbf{X}_{\psi,d}-\mathbf{X}_{\psi})
\\[6pt]
\ddot{\mathbf{p}}_{E,d}^{B}
+ K_{v,M} (\dot{\mathbf{p}}_{E,d}^{B}-\dot{\mathbf{p}}_{E}^{B})
+ K_{p,M} (\mathbf{p}_{E,d}^{B}-\mathbf{p}_{E}^{B})
\end{bmatrix},
\label{eq:outer_loop_pd}
\end{equation}

where $K_{p,\bullet}\succ 0$ and $K_{v,\bullet}\succ 0$ are diagonal 
gain matrices.\\
Due to the underactuated nature of the quadrotor base, translational motion cannot be directly commanded along arbitrary directions. 
In particular, the QP-based controller optimizes instantaneous task acceleration errors with respect to the current decision variables, but it does not inherently account for the need to reorient the vehicle in order to expand the admissible actuation space.
As a result, the solver may fail to generate base rotations that are necessary to realize the desired translational accelerations, even when such motions are dynamically feasible.

%------------------------------------------------------------
% Translational dynamics and thrust-direction constraint
%------------------------------------------------------------
The translational dynamics of the quadrotor base without the manipulator are given by
\begin{equation}
m_B\,\ddot{\mathbf{p}}_B
=
f_{z,b}\,R e_3
-
m_B g e_3 ,
\label{eq:trans_dyn}
\end{equation}
with $e_3 = [0\;0\;1]^\top$ denotes the body-fixed thrust axis.

Equation~\eqref{eq:trans_dyn} shows that translational motion can only be generated along the thrust direction
\begin{equation}
\mathbf{X}_t = R e_3 .
\label{eq:actual_thrust_dir}
\end{equation}

Given the desired translational acceleration $\ddot{\mathbf{p}}_B^{*}$ provided by the outer loop, the corresponding desired thrust direction is defined as~\cite{lee2010geometric}:
\begin{equation}
\mathbf{X}_{t,d}
=
\frac{m_B\left(\ddot{\mathbf{p}}_B^{*} + g e_3\right)}
{\left\|m_B\left(\ddot{\mathbf{p}}_B^{*} + g e_3\right)\right\|}.
\label{eq:desired_thrust_dir}
\end{equation}
where we assume that: 
\begin{equation}
    \left\|m_B\left(\ddot{\mathbf{p}}_B^{*} + g e_3\right)\right\| \neq 0
\end{equation}

Rather than enforcing full attitude tracking, the control objective is to align the actual thrust direction $X_t$ with its desired counterpart $X_{t,d}$, thereby ensuring that the vehicle reorients itself to make the desired translational acceleration achievable. To this end, a second-order PD law is imposed at the acceleration level:
\begin{equation}
\ddot{\mathbf{X}}_t^{*}
=
- \lambda_v \dot{X}_t
+ \lambda_p \left(X_{t,d} - X_t\right),
\label{eq:thrust_dir_pd}
\end{equation}
with $\lambda_p > 0$ and $\lambda_v > 0$ denoting proportional and damping gains.
%------------------------------------------------------------
% Augmented task acceleration
%------------------------------------------------------------
The resulting augmented desired task-acceleration vector is defined as
\begin{equation}
\ddot{\mathbf{X}}^{*}
=
\begin{bmatrix}
\ddot{\mathbf{p}}_B^{*} \\[6pt]
\ddot{\mathbf{X}}_{\psi}^{*} \\[6pt]
\ddot{\mathbf{p}}_{E}^{B*} \\[6pt]
\ddot{\mathbf{X}}_t^{*}
\end{bmatrix}
=
\begin{bmatrix}
\ddot{\mathbf{p}}_{B,d}
+ K_{v,B}(\dot{\mathbf{p}}_{B,d}-\dot{\mathbf{p}}_B)
+ K_{p,B}(\mathbf{p}_{B,d}-\mathbf{p}_B)
\\[6pt]
\ddot{\mathbf{X}}_{\psi,d}
+ K_{v,\psi}(\dot{\mathbf{X}}_{\psi,d}-\dot{\mathbf{X}}_{\psi})
+ K_{p,\psi}(\mathbf{X}_{\psi,d}-\mathbf{X}_{\psi})
\\[6pt]
\ddot{\mathbf{p}}_{E,d}^{B}
+ K_{v,M}(\dot{\mathbf{p}}_{E,d}^{B}-\dot{\mathbf{p}}_{E}^{B})
+ K_{p,M}(\mathbf{p}_{E,d}^{B}-\mathbf{p}_{E}^{B})
\\[6pt]
- \lambda_v \dot{X}_t
+ \lambda_p \left(X_{t,d}-X_t\right)
\end{bmatrix}.
\label{eq:new_task_acceleration}
\end{equation}

\subsection{QP-Based Optimization Controller}

The inner loop computes feasible generalized accelerations
$\dot{\boldsymbol{\nu}}\in\mathbb{R}^{6+n}$ that reproduce t
he desired task acceleration $\ddot{X}^*$ while satisfying all kinematic, 
dynamic, constraints by solving the following 
constrained quadratic optimization problem:
\begin{equation}
\begin{aligned}
\dot{\boldsymbol{\nu}}_{r}
&=\underset{\dot{\boldsymbol{\nu}}}{\arg\min}\;
\frac{1}{2}\Big(
\|\ddot{\mathbf{X}} - \ddot{\mathbf{X}}^{*}\|_{W_1}^{2}
+
\|\dot{\boldsymbol{\nu}} + \lambda \boldsymbol{\nu}\|_{W_2}^{2}
\Big)
\\[3pt]
\text{s.t.}\qquad
&A_{\mathrm{eq}}\dot{\boldsymbol{\nu}} = b_{\mathrm{eq}},\qquad
A\dot{\boldsymbol{\nu}} \le b,
\end{aligned}
\label{eq:joints_qp}
\end{equation}
where the task acceleration $\ddot{X}$ is given by
\begin{equation}
\ddot{X}
=
\begin{bmatrix}
\mathbf{I}_3 & \mathbf{0}_3 & \mathbf{0}_3 \\[4pt]
\mathbf{0}_3 & S(\mathbf{X}_{\psi}) & \mathbf{0}_3 \\[4pt]
\mathbf{0}_3 & \mathbf{0}_3 & J_{E}^{B}\\[4pt]
\mathbf{0}_3 & -R\,S(e_3) & \mathbf{0}_3 
\end{bmatrix}
\dot{\boldsymbol{\nu}}
+
\begin{bmatrix}
\mathbf{0}_3 \\[4pt]
S^{2}(\omega_{B})\,\mathbf{X}_{\psi} \\[4pt]
\dot{J}_{E}^{B}\,\mathbf{\dot{q}}_M\\[4pt]
 R\,S^{2}(\omega_{B})\,e_3
\end{bmatrix},
\end{equation}
with $J_{E}^{B}\in\mathbb{R}^{3\times n}$ is the
 translational Jacobian of the end effector expressed
  in $\mathcal{F}_B$.

The regularization term $\|\dot{\boldsymbol{\nu}} + \lambda 
\boldsymbol{\nu}\|_{W_2}^2$ 
penalizes excessive joint accelerations and velocities, 
ensuring smoothness of the commanded motion, and robustness to singularities.\\

The weighting matrices $W_1$, $W_2$, 
and the damping coefficient $\lambda$ are positive 
definite and tune the balance between task 
tracking and control regularity.  
All constraint definitions are detailed in the following subsection.

We adopt an inverse–dynamics control law to compute the reference input for~\eqref{eq:gauss_dynamics}.  
Given the optimal generalized acceleration $\dot{\boldsymbol{\nu}}_r$, the feedforward torque is
\begin{equation}
\mathbf{u}_{p}
=
M(\mathbf{q})\,\dot{\boldsymbol{\nu}}_r
+
C(\mathbf{q},\boldsymbol{\nu})\,\boldsymbol{\nu}
+
G(\mathbf{q}),
\label{eq:control_inputs}
\end{equation}
which is subsequently augmented with a passivity–based integral action to improve robustness.

\subsection{Passivity-Based Integral Action}

Following the formulations in~\cite{cisneros2018robust} and the passivity-based methods used in~\cite{garofalo2021adaptive, landau1989applications}, the control input is modified as
\begin{equation}
\mathbf{u} = \mathbf{u}_{p} + L\,\mathbf{s},
\label{new_control_input}
\end{equation}
where $\mathbf{s} = \boldsymbol{\nu}_{r} - \boldsymbol{\nu}$ is the velocity error, and  
$L \in \mathbb{R}^{(6+n)\times(6+n)}$ is the integral gain matrix.  
The reference velocity is obtained by integrating the optimal acceleration:
\begin{equation}
\boldsymbol{\nu}_{r}(t)
=
\int_{t_0}^{t} \dot{\boldsymbol{\nu}}_r(\tau)\, d\tau.
\end{equation}

Using the passivity-based design principle of~\cite{garofalo2020hierarchical}, the integral gain is selected as
\begin{equation}
L = C(\mathbf{q},\boldsymbol{\nu}) + K,
\label{eq:passivity_L}
\end{equation}
where $K$ is a symmetric positive-definite matrix.

Substituting~\eqref{eq:passivity_L} into~\eqref{eq:control_inputs} gives the final control input:
\begin{equation}
\mathbf{u}
=
M(\mathbf{q})\,\dot{\boldsymbol{\nu}}_{r}
+
C(\mathbf{q},\boldsymbol{\nu})\,\boldsymbol{\nu}_{r}
+
G(\mathbf{q})
+
K\,\mathbf{s}.
\label{eq:passivity_dynamics}
\end{equation}
This control law guarantees exponential convergence of the velocity error \(\mathbf{s}\), as demonstrated in~\cite{cisneros2018robust}.  
For practical implementation, physical and actuation constraints are incorporated in the QP problem, as described next.

\subsection{Constraints}

The optimization problem is subject to the following constraints.

\subsubsection{Underactuation constraint}

The quadrotor cannot generate control forces along its body $x$- and $y$-axes.  
Thus, the total wrench must satisfy
\begin{equation}
\begin{bmatrix} \mathbf{I}_2 & \mathbf{0}_{2\times1} \end{bmatrix}
\begin{bmatrix} R^\top & \mathbf{0}_{3\times(3+n)} \end{bmatrix}
\big( M\dot{\boldsymbol{\nu}} + C\boldsymbol{\nu} + G + L\mathbf{s} \big)
= 0.
\end{equation}
This can be expressed in linear form as
\begin{equation}
\begin{cases}
A_{\mathrm{eq}}
=
\begin{bmatrix}
\mathbf{I}_2 & \mathbf{0}_{2\times1}
\end{bmatrix}
\!
\begin{bmatrix}
R^\top & \mathbf{0}_{3\times(3+n)}
\end{bmatrix}
M,
\\[4pt]
b_{\mathrm{eq}}
=
-\begin{bmatrix}
\mathbf{I}_2 & \mathbf{0}_{2\times1}
\end{bmatrix}
\!
\begin{bmatrix}
R^\top & \mathbf{0}_{3\times(3+n)}
\end{bmatrix}
\big( C\boldsymbol{\nu} + G + L\mathbf{s} \big),
\end{cases}
\end{equation}
ensuring that only body–$z$ thrust and torques contribute to actuation.

\subsubsection{Actuation limits}
The individual rotor thrusts are obtained through the wrench–allocation model~\eqref{eq:wrench_allocation} and must satisfy the admissible thrust range of each rotor:

\begin{equation}
    F_{\min} \;\le\; \Lambda_{1}\, f_{z,b} + \Lambda_{2}\, \tau_b \;\le\; F_{\max},
    \label{eq:thrust_box}
\end{equation}
where $\Lambda_{1}\in\mathbb{R}^{4\times 1}$ and $\Lambda_{2}\in\mathbb{R}^{4\times 3}$ are the column partitions of
$\Xi^{-1}\in\mathbb{R}^{4\times 4}$.
Projecting~\eqref{eq:passivity_dynamics} along the body–$z$ axis yields
\begin{equation}
f_{z,b}
=
\begin{bmatrix}
e_3^\top R^\top & \mathbf{0}_{1\times(3+n)}
\end{bmatrix}
\big(
M\dot{\boldsymbol{\nu}}
+
C\boldsymbol{\nu}
+
G
+
L\mathbf{s}
\big),
\label{eq:thrust_expression}
\end{equation}

The body torques follow from the dynamic model and are written as
\begin{equation}
    \tau_{b}
    =
    \begin{bmatrix}
        \mathbf{0}_{3\times 3} &
        \mathbf{I}_{3} &
        \mathbf{0}_{3\times(3+n)}
    \end{bmatrix}
    \left( M\,\dot{\boldsymbol{\nu}} + C\,\boldsymbol{\nu} + G + L\,\mathbf{s} \right).
    \label{eq:torque_expression}
\end{equation}

Substituting~\eqref{eq:thrust_expression}–\eqref{eq:torque_expression} on \eqref{eq:thrust_box} yields the compact inequality
\begin{equation}
\begin{cases}
    \mathbf{A}_{F,\max}\, \dot{\boldsymbol{\nu}} \le \mathbf{b}_{F,\max}, \\[4pt]
    \mathbf{A}_{F,\min}\, \dot{\boldsymbol{\nu}} \le \mathbf{b}_{F,\min}.
\end{cases}
\end{equation}

\subsubsection{Joint position limits}

Each manipulator joint is constrained to remain within its mechanical range. Acceleration bounds are computed using a Grönwall-type inequality to ensure that predicted joint trajectories remain strictly inside the position limits~\cite{rossi2016trajectory}. Additional constraints such as collision avoidance and workspace boundaries can be incorporated in the same manner. All inequality constraints are finally collected in the standard form
\[
    A\,\dot{\boldsymbol{\nu}} \le b.
\]

\section{Simulation Results}

The proposed controller is validated on a lightweight aerial manipulator composed of a quadrotor equipped with a 5-DoF robotic arm, as shown in Fig.~\ref{fig:system_frames}.  
The quadrotor mass is \(m = 1.5~\mathrm{kg}\), and the body inertia matrix is
\(
J_b = \mathrm{diag}(0.01826,\,0.01826,\,0.03512)\,\mathrm{kg\,m^2}.
\)
The link masses, centers of mass, and inertia tensors of 
the manipulator are provided in the public repository 
(tag \texttt{v1.0}), together with the complete CAD/STEP model.
\footnote{
The simulation code and models are available at:
\url{https://github.com/nessrine-laribi/underactuated-aerial-manipulator-control}.
}

All simulations are carried out in MATLAB\textsuperscript{\tiny\textregistered} using Simscape Multibody\texttrademark. The physical parameters and actuator limits used in simulation are summarized in Table~\ref{tab:physical_params}.

\begin{table}[h]
    \centering
    \caption{Physical parameters and actuator limits of the simulated system.}
    \renewcommand{\arraystretch}{1.15}
    \begin{tabular}{|l|}
        \hline
        \textbf{Parameter}  \\ \hline
        Rotor thrust limits \([F_{\min},\,F_{\max}]\) [N]: \([0,\,15]\) \\
        Joint position limits \(q_{\max}\) [rad]: \([\,\pi,\, \pi/2,\, \pi/4,\, \pi,\, \pi/3\,]\) \\
        Joint position limits \(q_{\min}\) [rad]: \(-[\,\pi,\, \pi/2,\, \pi/4,\, \pi,\, \pi/3\,]\) \\
        Joint torque limits \(\tau_{\max}\) [Nm]: \([1.6,\, 5,\, 5,\, 5,\, 5]\) \\
        Joint torque limits \(\tau_{\min}\) [Nm]: \(-[1.6,\, 5,\, 5,\, 5,\, 5]\) \\
        \hline
    \end{tabular}
    \label{tab:physical_params}
\end{table}

The controller gains and weight matrices are reported in Table~\ref{tab:control_gains}.  
The integral gain is chosen as \(K = 10\,M+0.001\,I_{11}\),
 where \(M\) denotes the system mass matrix.\\
The control loop runs at \(200~\mathrm{Hz}\), and the QP problem is solved using \texttt{quadprog}.  
System dynamics are integrated using \texttt{ode45} with a maximum step size of
\(5\times10^{-2}~\mathrm{s}\) and a relative tolerance of \(10^{-3}\).

\begin{table}[h]
    \centering
    \caption{Controller gains and optimization weights.}
    \renewcommand{\arraystretch}{1.15}
    \begin{tabular}{|c|c|c|c|}
        \hline
        \textbf{Parameter} & \textbf{Value} & \textbf{Parameter} & \textbf{Value} \\ \hline
        $K_{p,B}$ & $diag([4,\,4,\,9])$ & $K_{v,B}$ & $diag([4,\,4,\,6])$ \\
        $K_{p,\psi}$ & $9\,I_3$ & $K_{v,\psi}$ & $6\,I_3$  \\
        $K_{p,M}$ & $49\,I_3$ & $K_{v,M}$ & $14\,I_3$ \\
        $\lambda_{p}$ & $100\,I_3$ & $\lambda_{v}$ & $diag([10,\,10,\,20])$ \\
        $W_{1}$ & $blkdiag(100\,I_3,\,50\,I_3,\,500\,I_3,\,500\,I_3)$ &  &  \\
        $W_{2}$ & $blkdiag(100\,I_3,\,50\,I_3,\,0.01\,I_5)$ 
        & $\lambda$ & $blkdiag(1\,I_3,\,10\,I_3,\,100\,I_5)$ \\
        \hline
    \end{tabular}
    \label{tab:control_gains}
\end{table}

\begin{figure}[H]
    \centering
    \includegraphics[width=0.6\columnwidth]{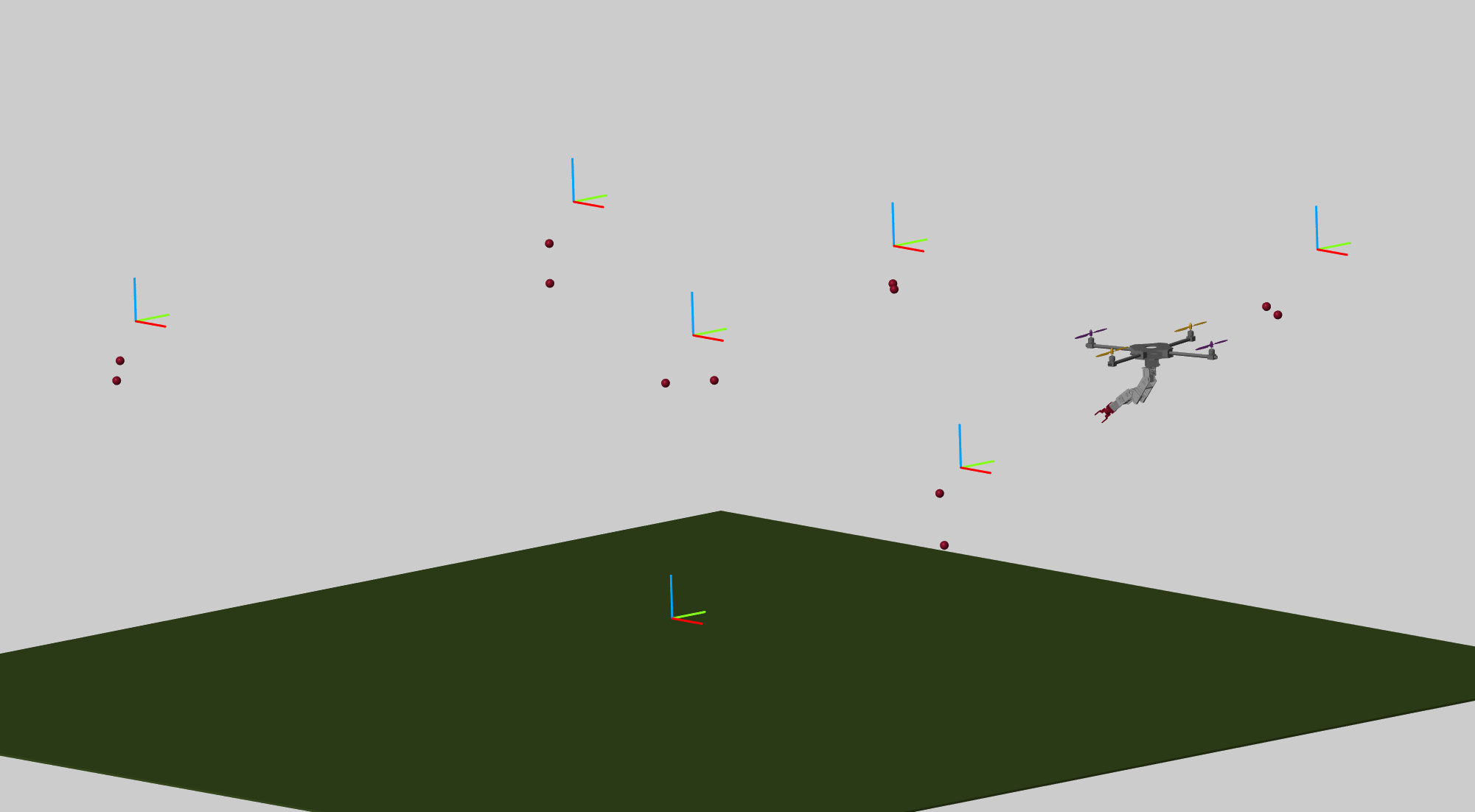}
    \caption{Simscape Multibody visualization of the aerial manipulator tracking the base position, yaw angle, and end-effector setpoints.}
    \label{fig:traj_screenshot}
\end{figure}

The aerial manipulator is commanded to track multiple setpoints for the floating–base position, desired yaw angle, and end-effector position, each generated along smooth polynomial trajectories with time-varying references (Fig.~\ref{fig:traj_screenshot}).

The robustness of the proposed controller is evaluated 
under model uncertainties by introducing a $-10\%$ mass 
perturbation on each manipulator link, neglecting the products 
of inertia in the controller model, and adding viscous joint 
damping of $5\times10^{-4}\,\mathrm{N\!\cdot\!m/(rad/s)}$.

\begin{figure}[h!]
  \centering
  \begin{subfigure}{0.50\columnwidth}
    \includegraphics[width=\columnwidth]{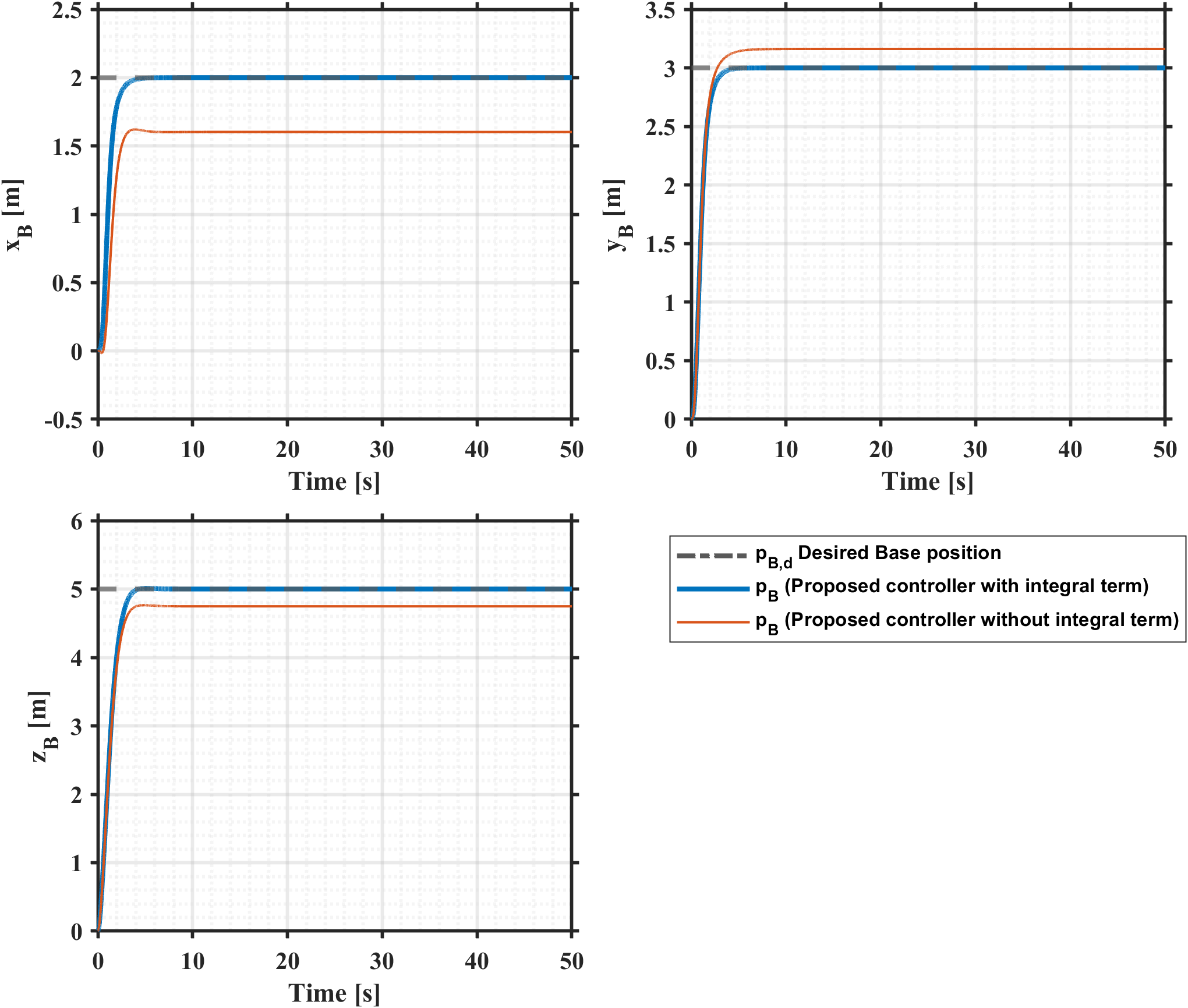}
    \caption{Base position regulation under set-point stabilization.}
    \label{quadcopter_pos_s1}
  \end{subfigure}
  \begin{subfigure}{0.45\columnwidth}
    \includegraphics[width=\columnwidth]{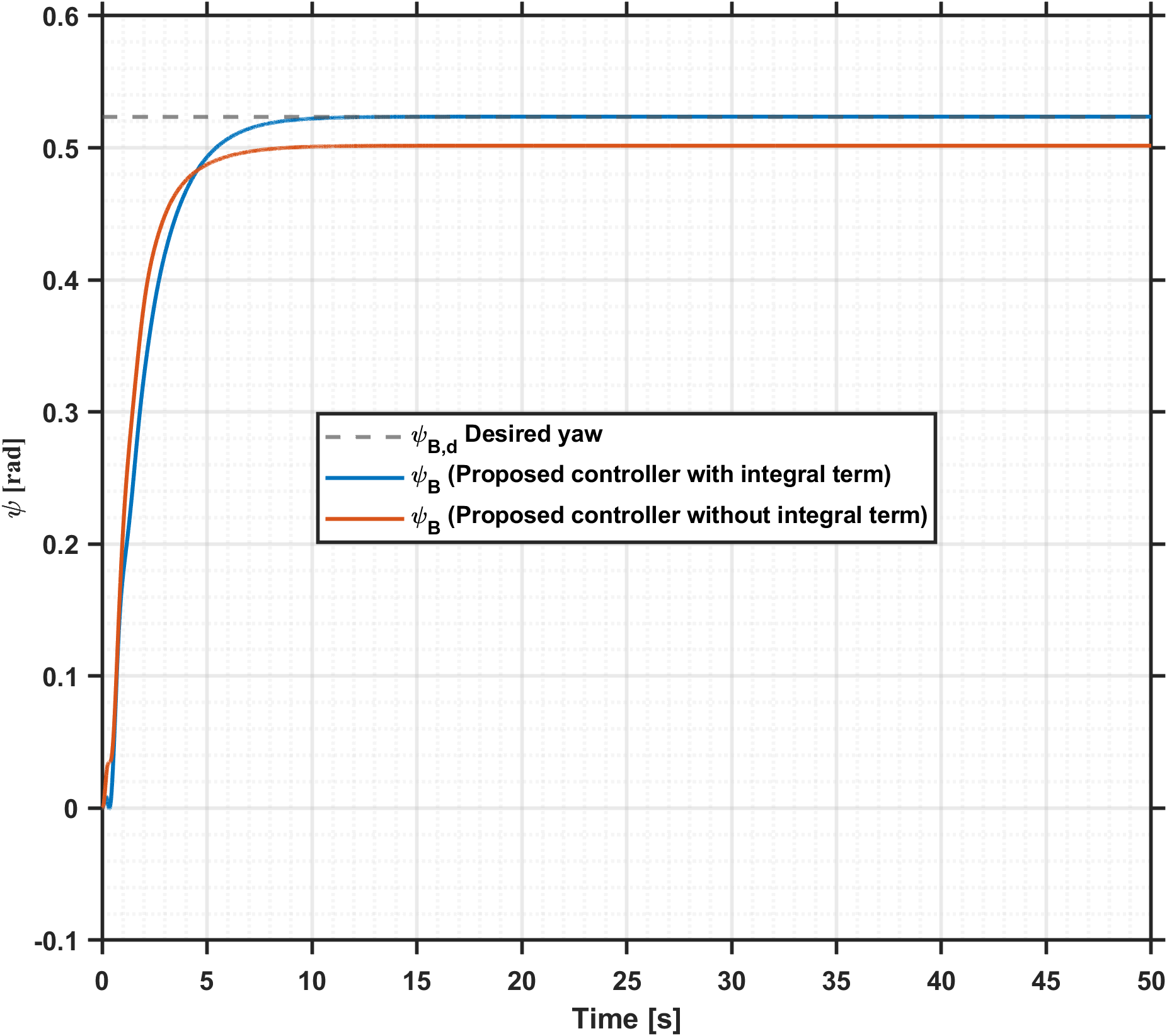}
    \caption{Yaw angle regulation under set-point stabilization.}
    \label{quadcopter_yaw_s1}
  \end{subfigure}
 \caption{Floating-base position and yaw angle regulation under set-point stabilization and model uncertainties.}
 \label{quadcopter_task_s1}
\end{figure}

\begin{figure}[h]
  \centering
    \includegraphics[width=0.65\columnwidth]{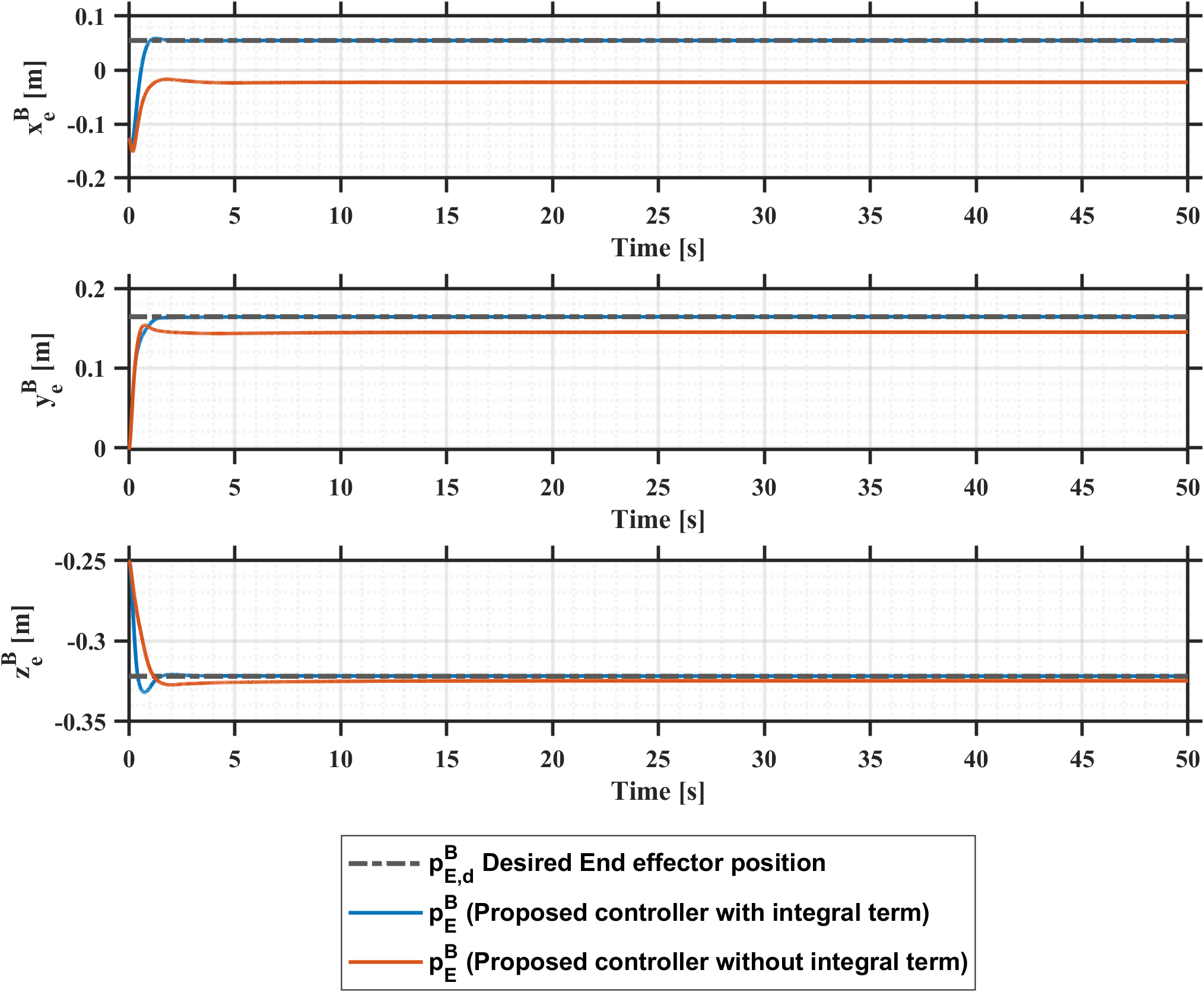}
   \caption{End-effector position regulation under set-point stabilization and model uncertainties.}
  \label{end_effector_task_s1}
\end{figure}

The evaluation starts with a set-point stabilization scenario. 
In the absence of integral action, the controller maintains 
bounded responses for both the floating base and the end-effector; 
however, small but persistent steady-state offsets are observed, 
particularly in the base position and yaw responses, 
as illustrated in Figs.~\ref{quadcopter_task_s1} 
and~\ref{end_effector_task_s1}. 
When the passivity-based integral term is included, 
these steady-state errors are effectively eliminated, 
yielding zero steady-state tracking error for 
both the floating base and the end-effector. 
This control law guarantees exponential convergence of 
the velocity error $\mathbf{s}$, 
and the simulation results confirm the convergence and 
stability properties established in~\cite{cisneros2018robust}.

\begin{figure}[h!]
  \centering
  \begin{subfigure}{0.52\columnwidth}
    \includegraphics[width=\columnwidth]{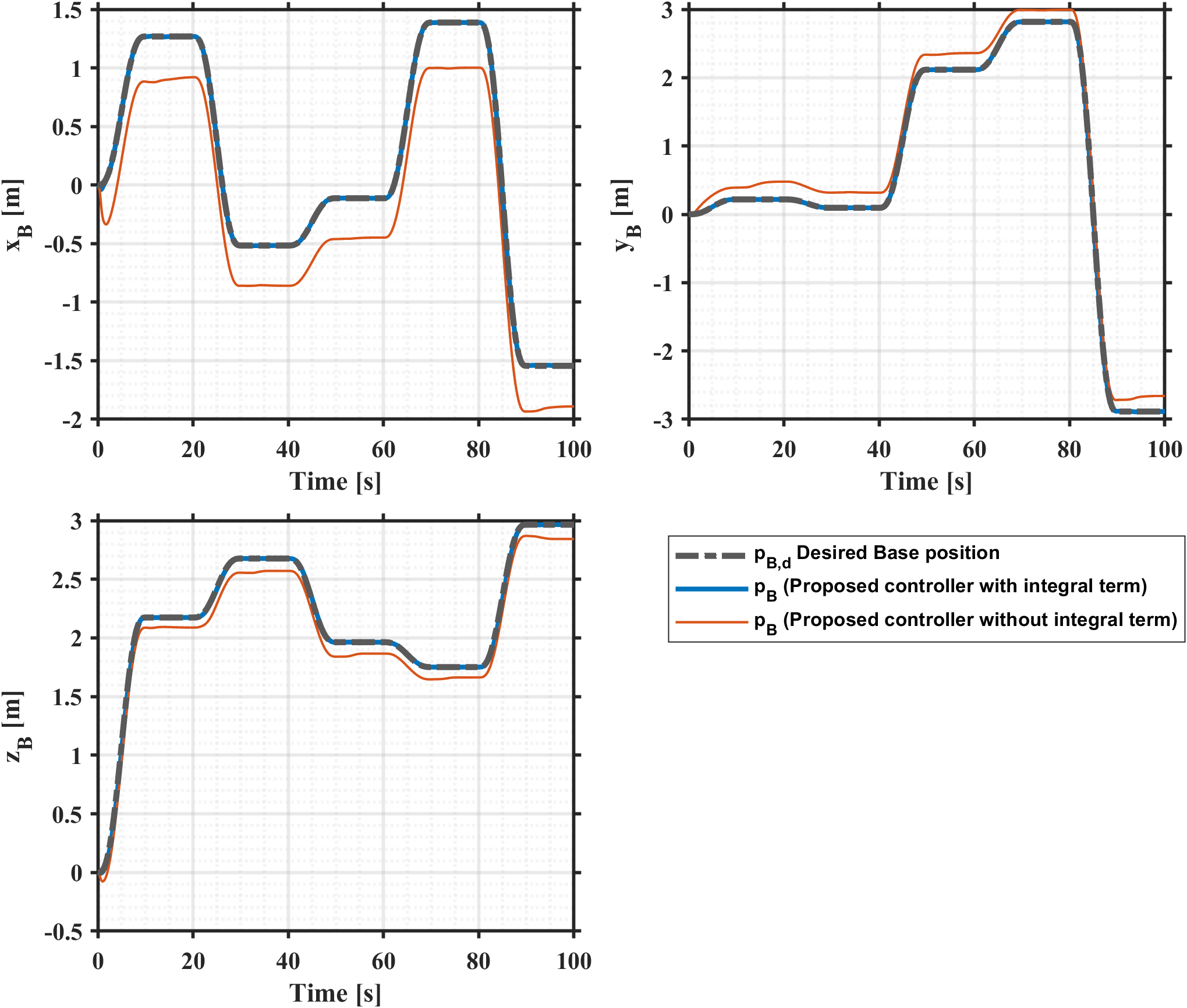}
    \caption{Base position tracking.}
    \label{quadcopter_pos}
  \end{subfigure}
  \begin{subfigure}{0.47\columnwidth}
    \includegraphics[width=\columnwidth]{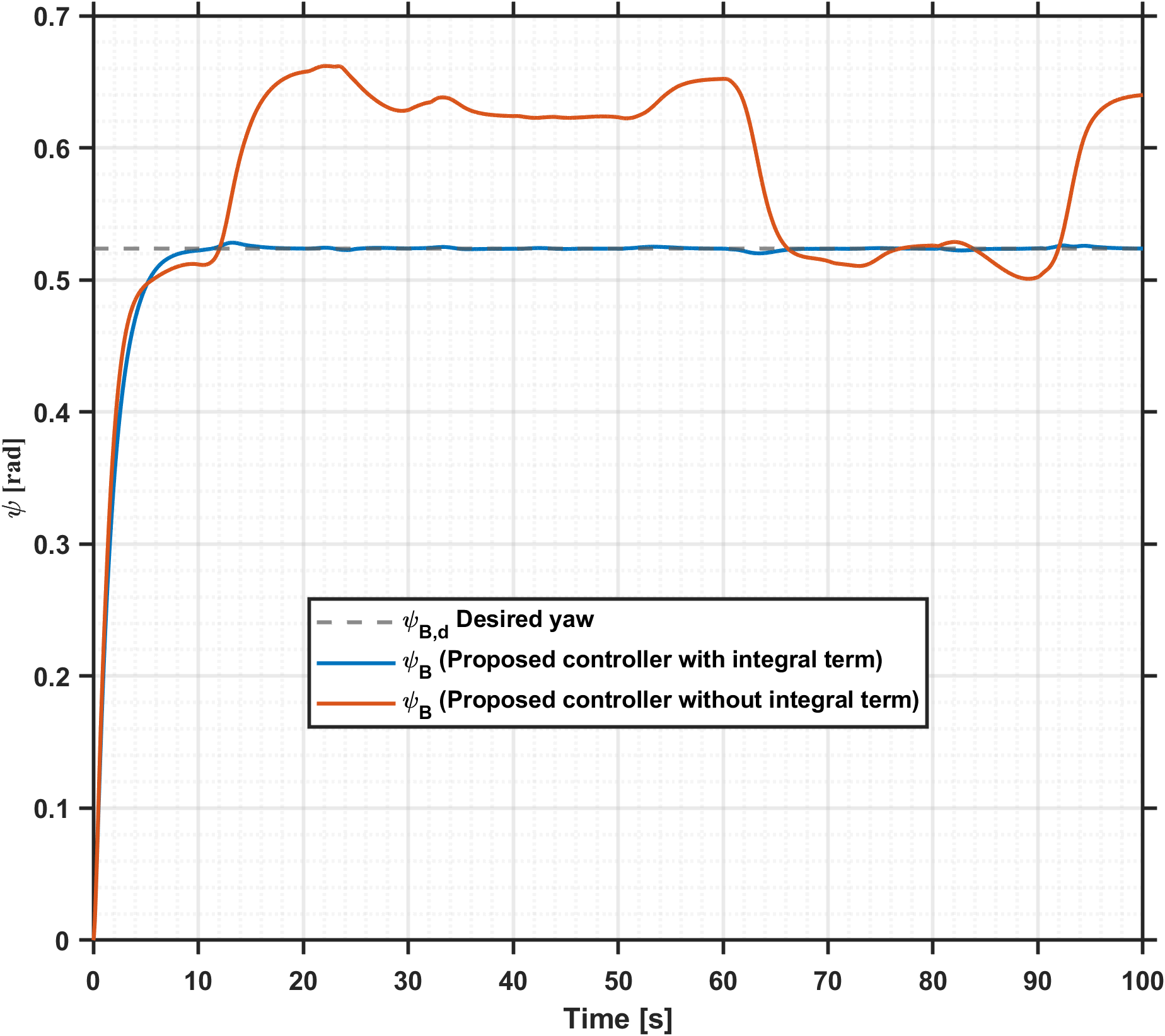}
    \caption{Yaw angle tracking.}
    \label{quadcopter_yaw}
  \end{subfigure}
 \caption{Floating-base position and yaw angle tracking under model uncertainties.}
\label{quadcopter_task}
\end{figure}

\begin{figure}[h]
  \centering
    \includegraphics[width=0.6\columnwidth]{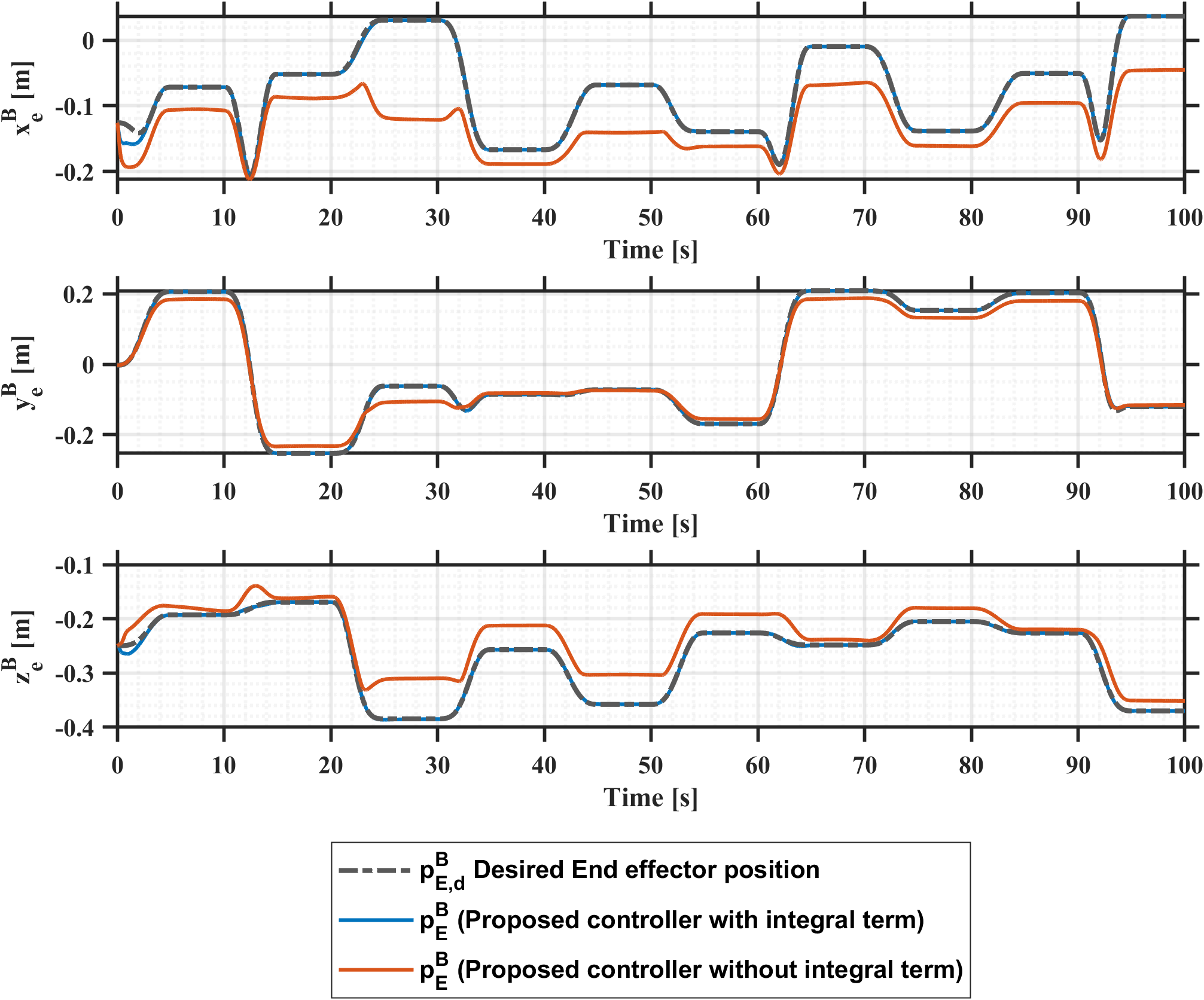}
   \caption{End-effector position tracking under model uncertainties.}
  \label{end_effector_task}
\end{figure}

Figures~\ref{quadcopter_pos},~\ref{quadcopter_yaw}, and~\ref{end_effector_task} report the results obtained for trajectory tracking with multiple set-points under the same uncertainty conditions. Without integral compensation, the floating-base position exhibits noticeable steady-state deviations, resulting in a root-mean-square error of
\(
\mathrm{RMSE}_{p,B} = [0.3635\;\;0.2058\;\;0.1075]~\mathrm{m}.
\)
Yaw tracking remains bounded, with a mean error of 
$0.0509~\mathrm{rad}$, 
although persistent offsets are visible along constant 
reference segments. 

The end-effector position relative to the base also shows reduced tracking accuracy, with
\(
\mathrm{RMSE}_{p_E^B} = [0.0609\;\;0.0200\;\;0.0355]~\mathrm{m}.
\) With the passivity-based integral term included, steady-state errors induced by model inaccuracies are effectively compensated. The base position and yaw responses closely follow the desired trajectories, yielding significantly reduced tracking errors,
\(
\mathrm{RMSE}_{p,B} = [0.0038\;\;0.0017\;\;0.0118]~\mathrm{m},
\)
and a mean yaw error of $0.0106~\mathrm{rad}$. A substantial improvement is also obtained for the end-effector tracking task, with errors decreasing by nearly one order of magnitude,
\(
\mathrm{RMSE}_{p_E^B} = [0.0038\;\;0.0017\;\;0.0022]~\mathrm{m}.
\)
This behavior highlights the effectiveness of the proposed passivity-based integral action in rejecting disturbances.

In all tested scenarios, the QP optimization problem remains feasible and converges within an average of three iterations per control step, confirming the real-time suitability of the proposed control architecture.

\begin{figure}[h]
  \centering
  \begin{subfigure}{0.45\columnwidth}
    \includegraphics[width=\columnwidth]{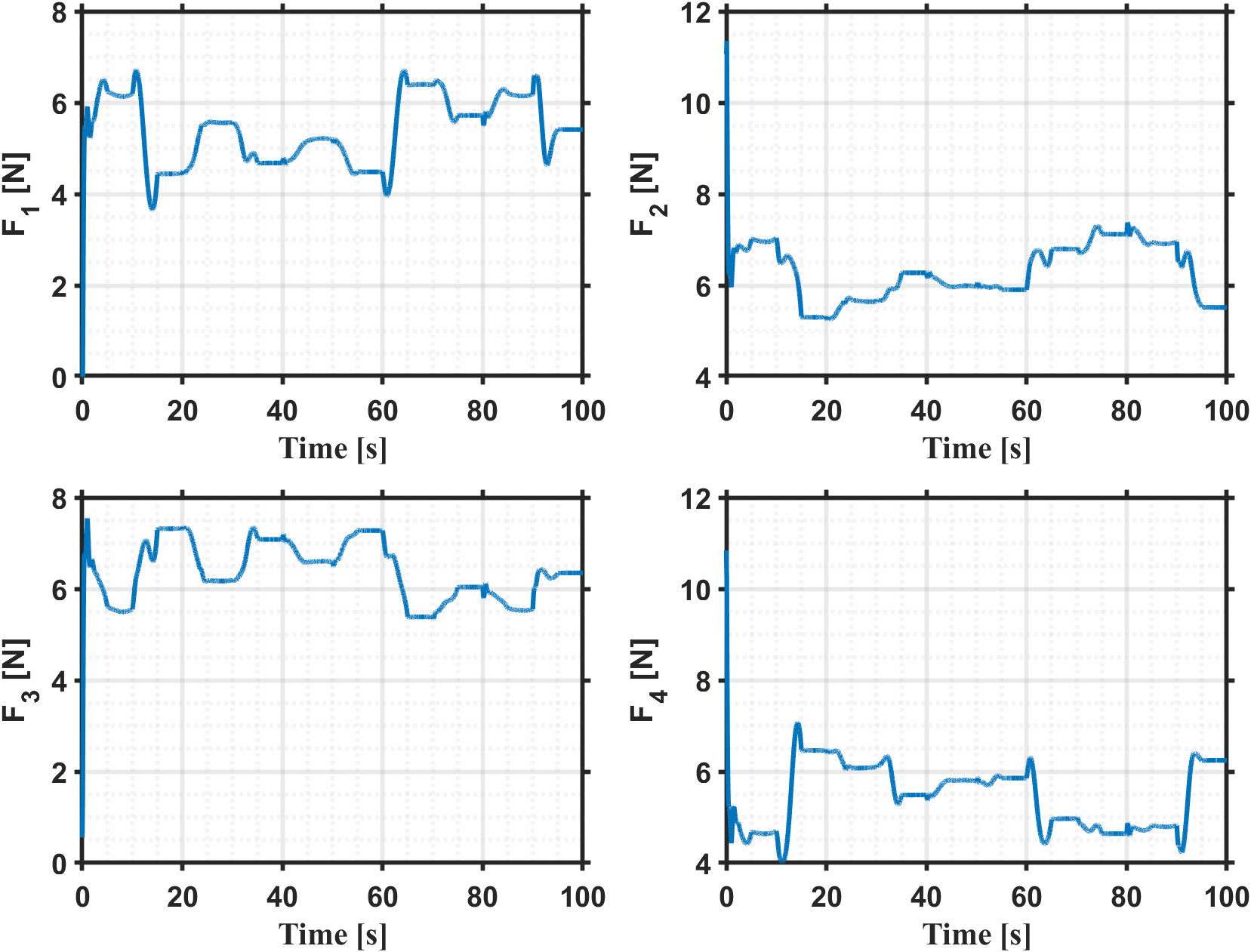}
    \caption{Rotor thrusts.}
  \end{subfigure}\hfill
  \begin{subfigure}{0.45\columnwidth}
    \includegraphics[width=\columnwidth]{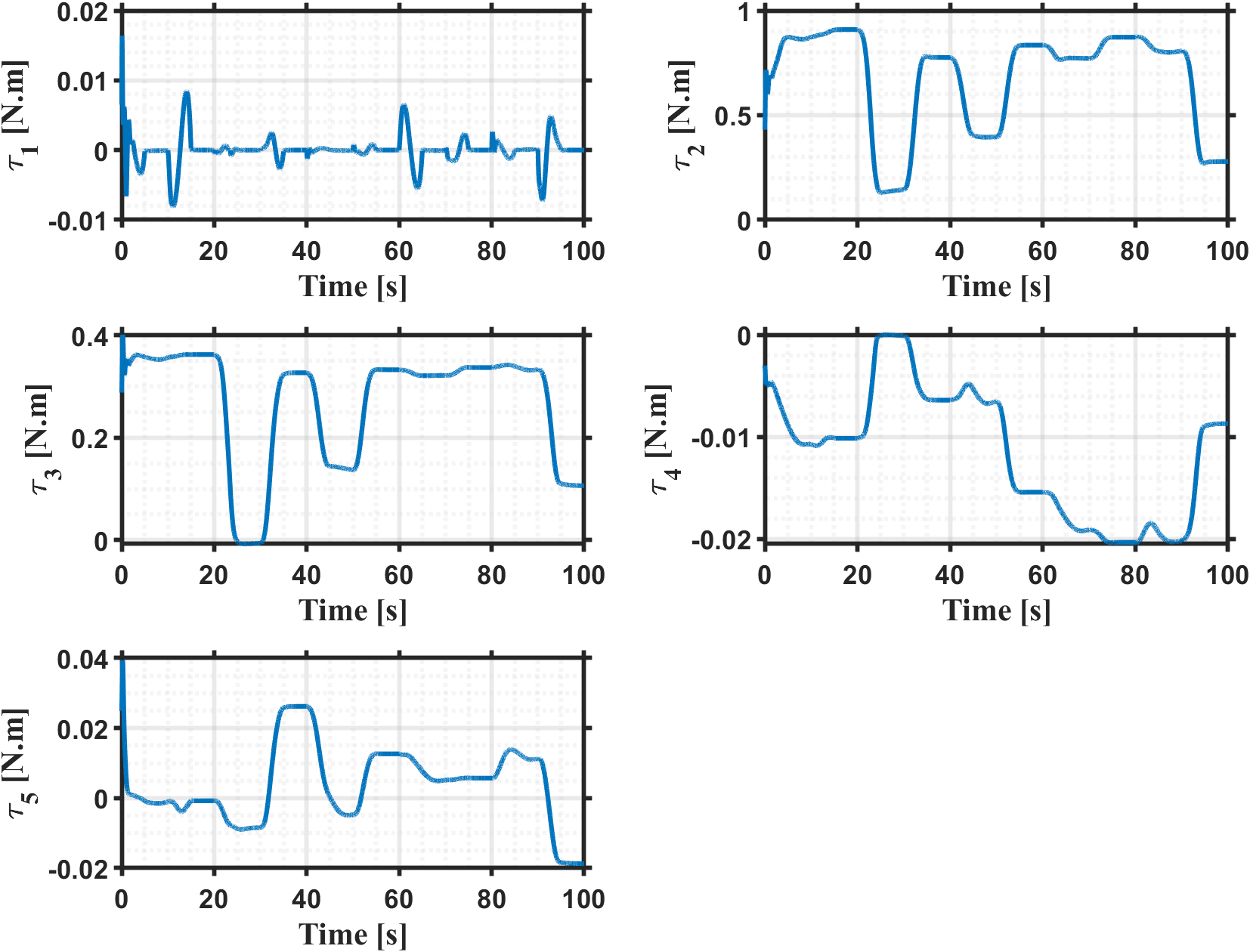}
    \caption{Joint torques.}
  \end{subfigure}
  \caption{Commanded rotor thrusts and joint torques within actuator limits.}
  \label{fig:s2_inputs}
\end{figure}

\begin{figure}[h!]
  \centering
  \includegraphics[width=0.65\columnwidth]{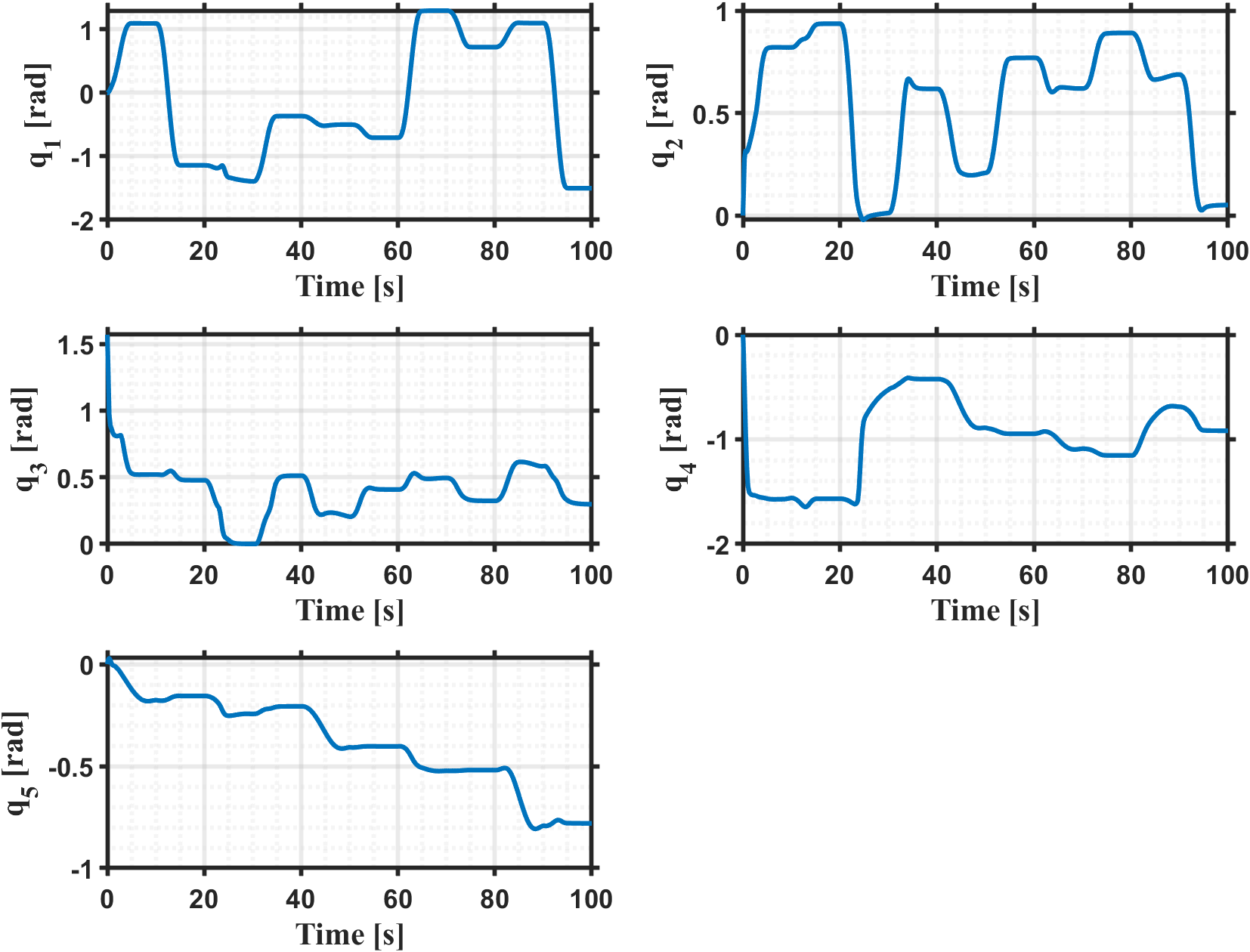}
  \caption{Manipulator joint trajectories within joint-space limits.}
  \label{fig:joint_limits}
\end{figure}

Figure~\ref{fig:s2_inputs} shows rotor thrusts \(F_{1:4}\) and joint torques \(\boldsymbol{\tau}\).  
All commands stay within the limits of Table~\ref{tab:physical_params}, confirming constraint satisfaction by the QP layer.  
Saturation events are brief and occur only during aggressive transients. Manipulator joint angles (Fig.~\ref{fig:joint_limits}) also remain within their bounds, verifying consistent enforcement of joint-space constraints.

To approximate realistic experimental conditions, ground-truth states were replaced with simulated sensor measurements generated using the UAV Toolbox. 
Quadrotor states are estimated from a simulated \textbf{RTK-GNSS} receiver (u-blox ZED-F9P) and a \textbf{6-axis IMU} (TDK ICM-42688-P). 
The GNSS operates at \(f_{\mathrm{GPS}} = 10~\mathrm{Hz}\) with \(1\sigma\) horizontal and vertical position accuracies of \(0.02~\mathrm{m}\), and a velocity accuracy of \(0.05~\mathrm{m\,s^{-1}}\). 
The IMU provides tri-axial specific force and angular rate measurements at \(200~\mathrm{Hz}\). 
Base position is estimated through a complementary filter, and attitude is reconstructed using Hua’s nonlinear observer~\cite{hua2010attitude}. 
Manipulator joint angles are measured with 12-bit absolute encoders (\(\Delta\theta = 2\pi/4096 \approx 1.53\times10^{-3}\,\mathrm{rad}\), quantization error \(\pm\Delta\theta/2\)), while joint velocities are obtained by finite differences followed by first-order low-pass filtering.

\begin{table}[H]
\centering
\caption{Simulated IMU parameters (TDK ICM-42688-P).}
\label{tab:imu_params}
\begin{tabular}{|c c c|}
\hline
Parameter & Value & Units/Notes \\
\hline
\multicolumn{3}{|c|}{\emph{Common}}\\
Rate $f_{\rm IMU}$ & 200 & Hz \\
\hline
\multicolumn{3}{|c|}{\emph{Accelerometer}}\\
Max reading & 156.9064 & m/s$^{2}$ \\
Resolution & 0.004787 & (m/s$^{2}$)/LSB  \\
Velocity random walk & $6.86\times10^{-4}$ & -- \\
Constant offset bias & {[}0\;0\;0{]} & m/s$^{2}$ \\
Bias instability & $1.0\times10^{-3}$ & m/s$^{2}$ \\
\hline
\multicolumn{3}{|c|}{\emph{Gyroscope}}\\
Max reading & 34.9066 & rad/s  \\
Resolution & 0.001065 & (rad/s)/LSB \\
Angle random walk & $4.89\times10^{-5}$ & (rad/s)/$\sqrt{\mathrm{Hz}}$ \\
Bias instability & $2.0\times10^{-4}$ & rad/s  \\
Constant offset bias & {[}0\;0\;0{]} & rad/s \\
\hline
\end{tabular}
\end{table}

\begin{figure}[h!]
  \centering
  \begin{subfigure}{0.52\columnwidth}
    \includegraphics[width=\columnwidth]{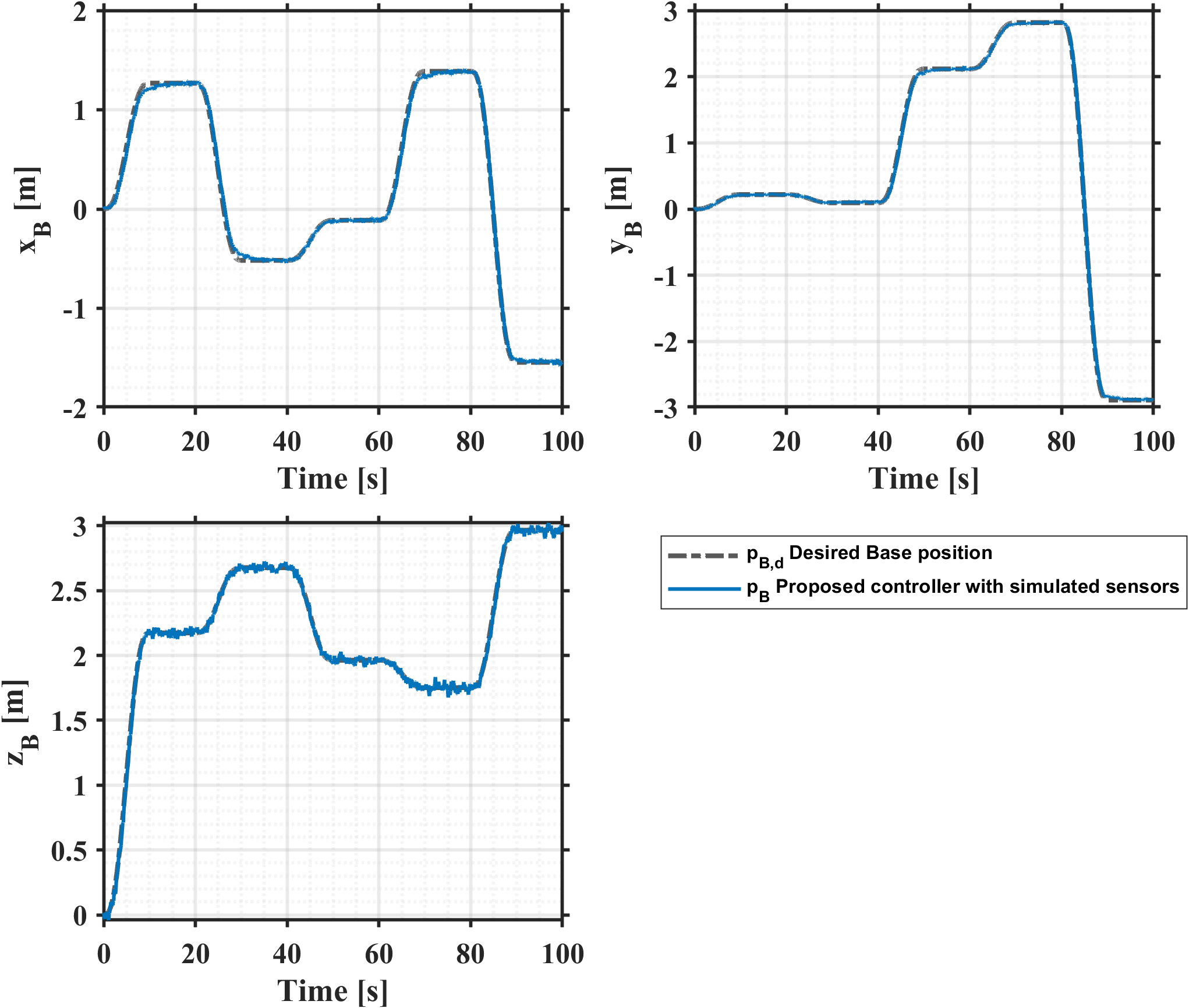}
    \caption{Base position tracking.}

  \end{subfigure}
  \begin{subfigure}{0.47\columnwidth}
    \includegraphics[width=\columnwidth]{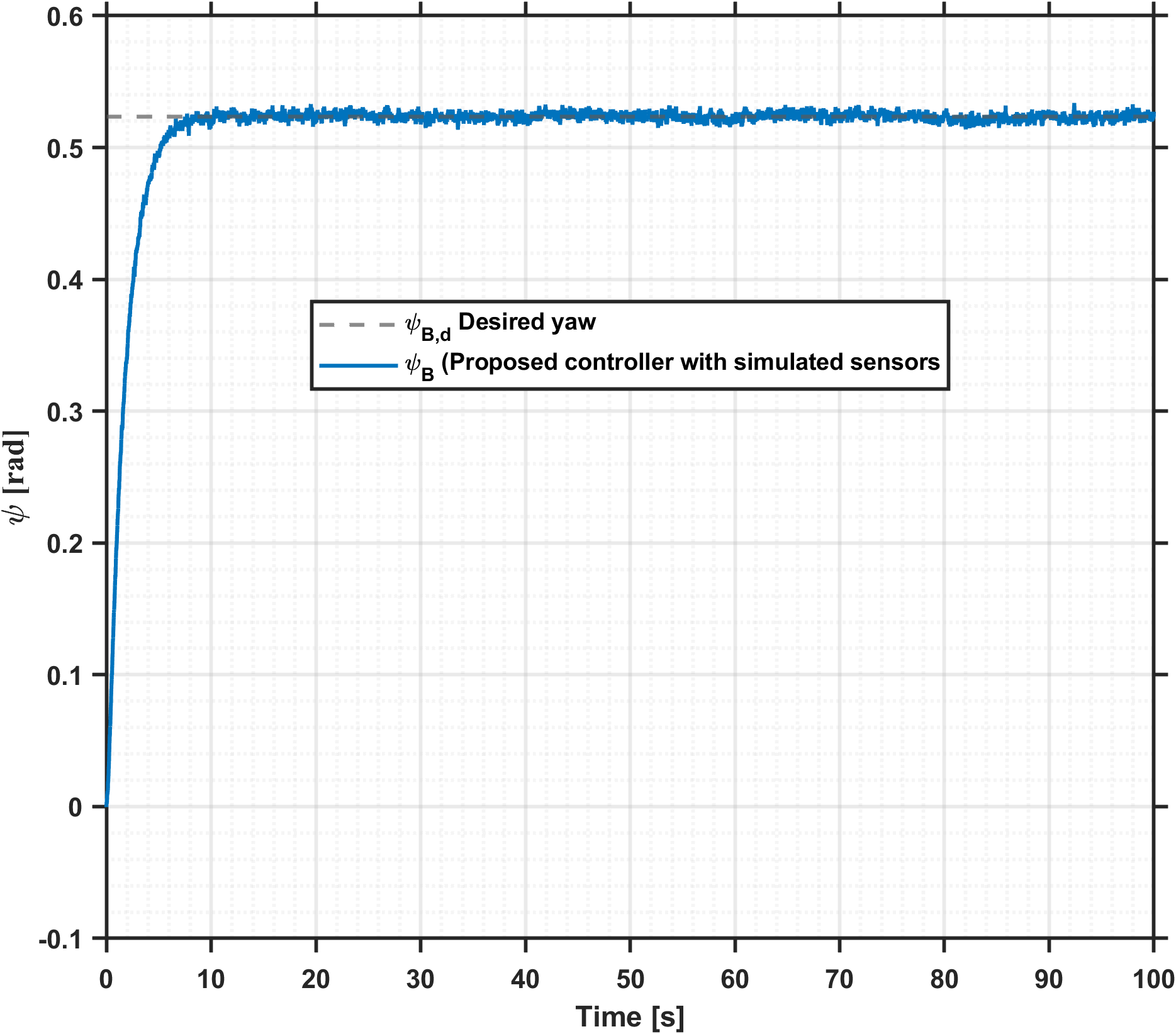}
   \caption{Yaw angle tracking.}
  \end{subfigure}
  \caption{floating base  position and yaw attitude tracking  using sensor-based state estimation}
  \label{fig:s3_track}
\end{figure}

\begin{figure}[h!]
  \centering
    \includegraphics[width=0.7\columnwidth]{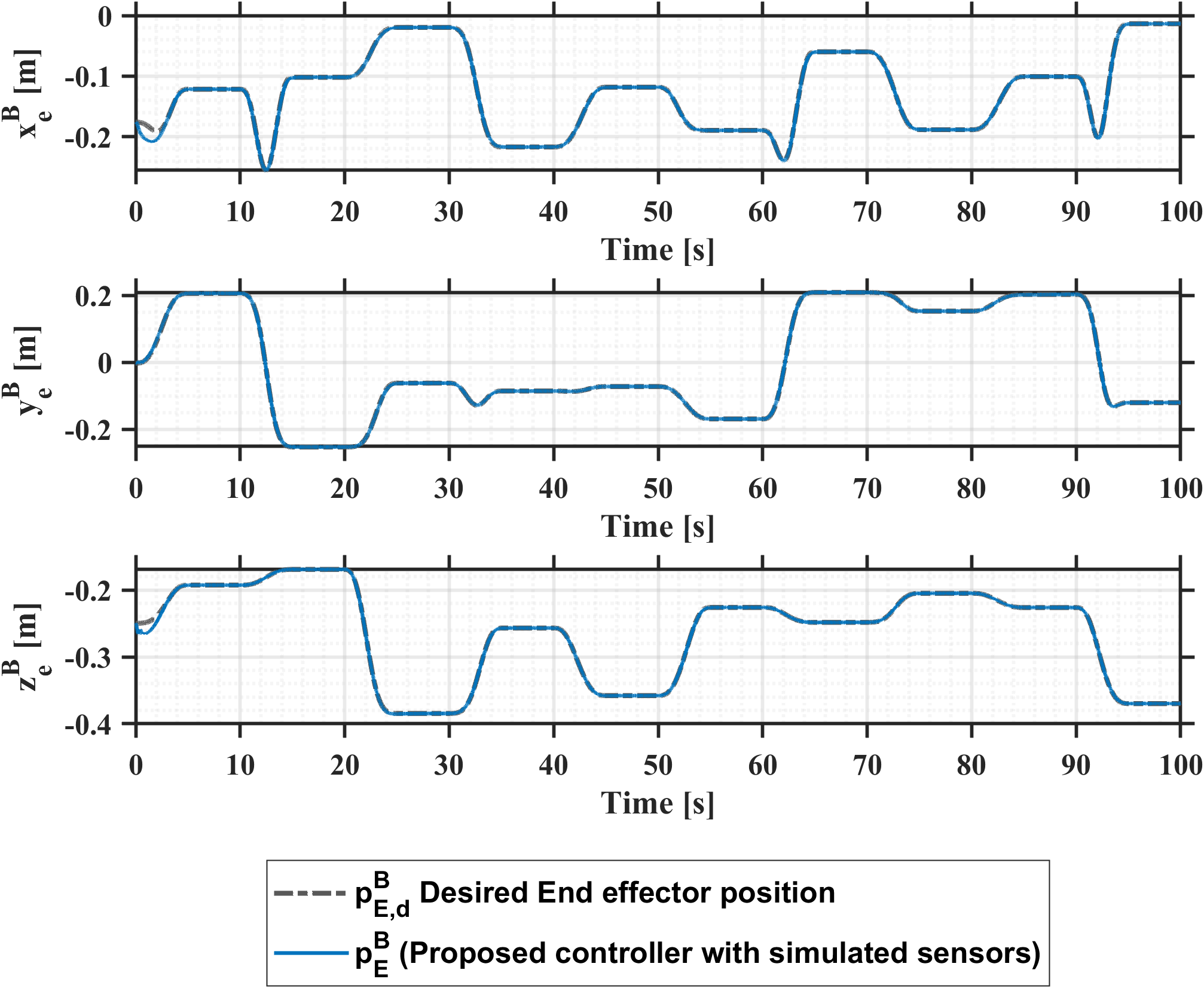}
    \caption{End-effector position tracking  using sensor-based state estimation}

  \label{fig:s3_track_effector}
\end{figure}

Figures~\ref{fig:s3_track}–\ref{fig:s3_track_effector} show that the proposed controller achieves accurate tracking of the floating-base states and the end-effector position using sensor-based state estimation.
Small steady-state errors and mild oscillations are observed near sharp reference transitions, mainly due to estimator dynamics and sensor noise.

\begin{figure}[h!]
  \centering
  \begin{subfigure}{0.5\columnwidth}
    \includegraphics[width=\columnwidth]{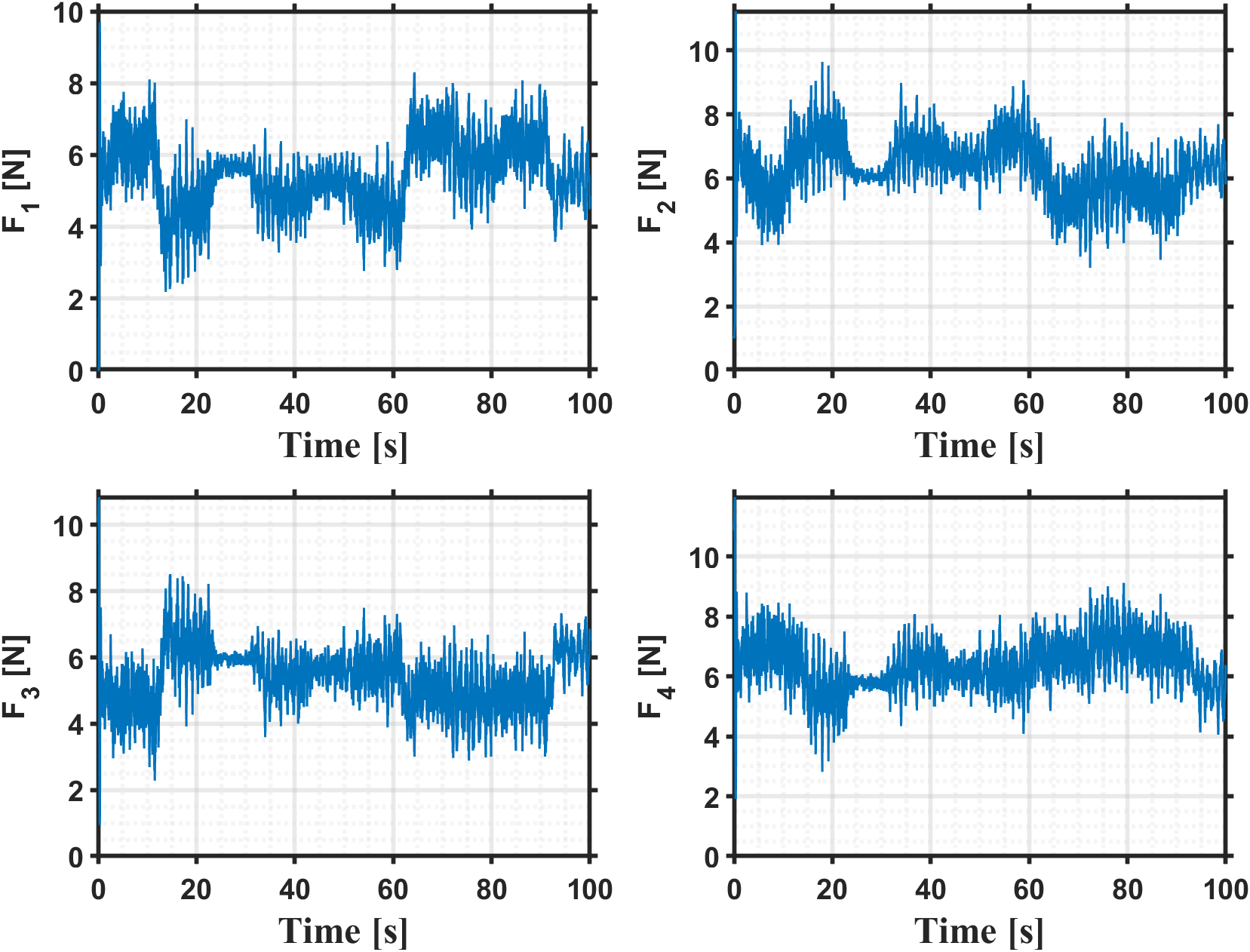}
    \caption{Rotor thrusts.}
      \label{fig:s3_thrust}
  \end{subfigure}\hfill
  \begin{subfigure}{0.5\columnwidth}
    \includegraphics[width=\columnwidth]{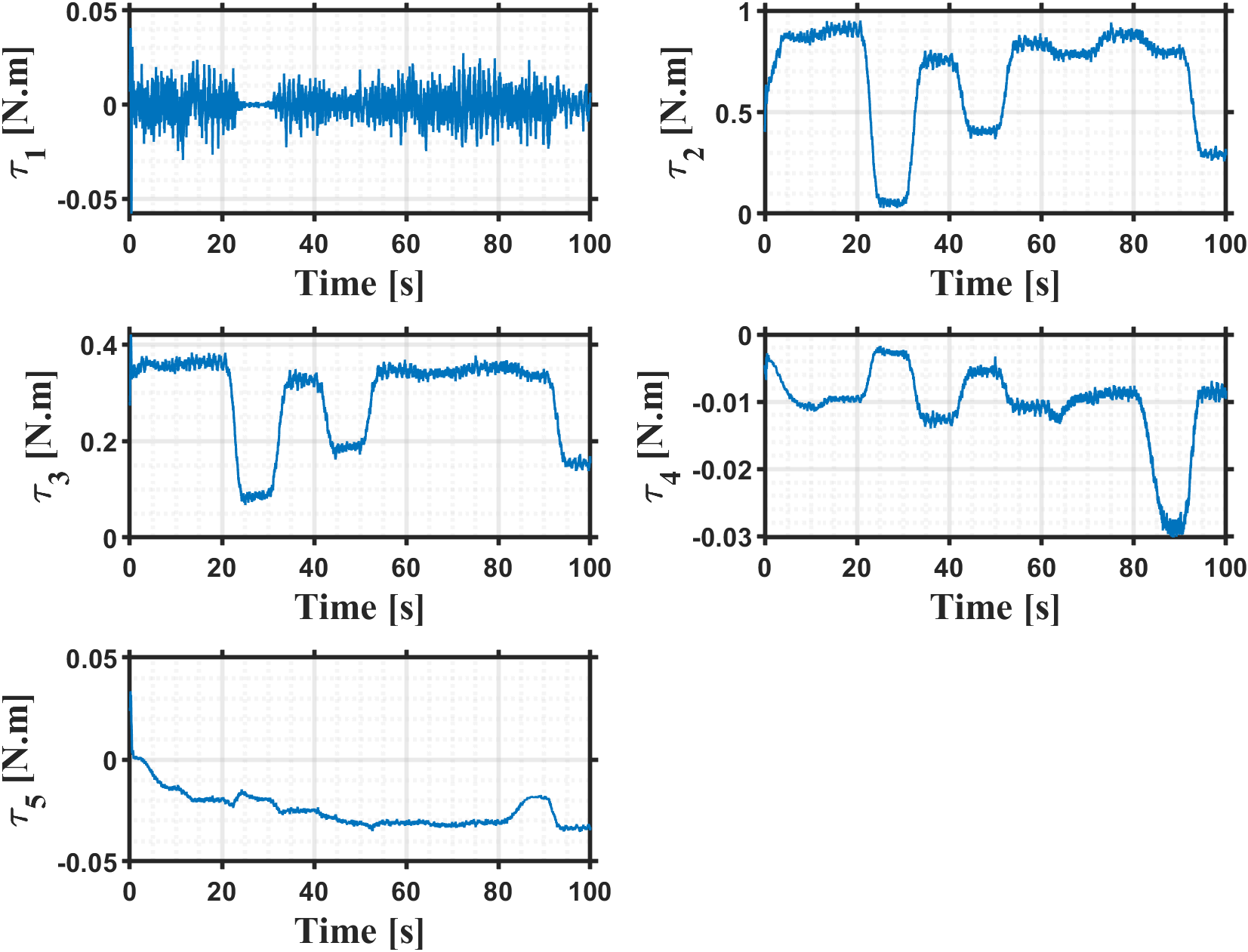}
    \caption{Joint torques.}
      \label{fig:s3_tau}
  \end{subfigure}
  \caption{Commanded rotor thrusts and joint torques within actuator limits.}
  \label{fig:s3_inputs}
\end{figure}

The corresponding actuator commands (Fig.~\ref{fig:s3_inputs}), including rotor thrusts (Fig.~\ref{fig:s3_thrust}) and manipulator joint torques (Fig.~\ref{fig:s3_tau}), remain within their physical limits, confirming safe operation.
Throughout the maneuver, the QP layer guarantees feasibility and smooth control allocation, demonstrating consistent coordination between tracking performance and actuator usage.
\section{Conclusion}

This work presented a coupled QP-based control framework for underactuated aerial manipulators, enabling accurate trajectory tracking through coordinated control of the aerial platform and the onboard manipulator. The proposed approach addresses the challenges arising from the underactuated nature of the quadrotor base by leveraging geometric control principles to guide the QP solver toward dynamically consistent solutions. In particular, the framework computes generalized accelerations that respect the system dynamics while explicitly enforcing joint limits, actuator constraints, and underactuation-induced feasibility conditions.

To enhance robustness against modeling uncertainties, sensor noise, and steady-state errors, a passivity-based integral action was incorporated at the torque level. High-fidelity simulations conducted in Simscape Multibody demonstrated accurate and stable tracking performance under realistic sensing and disturbance conditions. The proposed coupled strategy achieved improved steady-state accuracy, smoother control inputs, and reliable constraint satisfaction, while maintaining real-time feasibility of the QP solver.

Future work will focus on task-prioritized control formulations emphasizing direct end-effector motion control, together with a formal analysis of stability and convergence properties of the QP-based controller. Experimental validation on a physical aerial manipulator platform is also planned, along with extensions to contact-rich manipulation scenarios involving interaction forces and environmental constraints.

\bibliographystyle{elsarticle-num}
\bibliography{bibliography} 

@article{suarez2018design,
  title={Design and control of aerial manipulators for inspection and maintenance of power infrastructures},
  author={Suarez, A and Fagiano, L and Ryll, M and Franchi, A and others},
  journal={Annual Reviews in Control},
  volume={46},
  pages={161--176},
  year={2018},
  publisher={Elsevier}
}

@article{tognon2019truly,
  title={A truly-redundant aerial manipulator system with application to push-and-slide inspection in industrial plants},
  author={Tognon, Marco and others},
  journal={IEEE Robotics and Automation Letters},
  volume={4},
  number={2},
  pages={1846--1851},
  year={2019},
  publisher={IEEE}
}

@article{korpela2014towards,
  title={Towards valve turning using a dual-arm aerial manipulator},
  author={Korpela, Christopher and Orsag, Matko and Oh, Paul},
  journal={Journal of Intelligent \& Robotic Systems},
  volume={73},
  pages={361--376},
  year={2014},
  publisher={Springer}
}

@article{zhang2019robust,
  title={Robust control of an aerial manipulator based on a variable inertia parameters model},
  author={Zhang, Guangyu and others},
  journal={IEEE Transactions on Industrial Electronics},
  volume={67},
  number={11},
  pages={9515--9525},
  year={2019},
  publisher={IEEE}
}

@article{bulut2019decoupled,
  title={Decoupled Cascaded PID Control of an Aerial Manipulation System},
  author={Bulut, Nebi and others},
  journal={Hittite Journal of Science and Engineering},
  volume={6},
  number={4},
  pages={251--259},
  year={2019},
  publisher={Hitit University}
}

@inproceedings{lippiello2012exploiting,
  title={Exploiting redundancy in Cartesian impedance control of UAVs equipped with a robotic arm},
  author={Lippiello, Vincenzo and Ruggiero, Fabio},
  booktitle={2012 IEEE/RSJ International Conference on Intelligent Robots and Systems},
  pages={3768--3773},
  year={2012},
  organization={IEEE}
}

@inproceedings{jeong2023passivity,
  title={Passivity-based decentralized control for collaborative grasping of under-actuated aerial manipulators},
  author={Jeong, Jinyeong and Kim, Min Jun},
  booktitle={2023 IEEE International Conference on Robotics and Automation (ICRA)},
  pages={7699--7705},
  year={2023},
  organization={IEEE}
}

@article{nava2020direct,
  title={Direct force feedback control and optimization for an aerial manipulator},
  author={Nava, Giulio and others},
  journal={Robotics and Autonomous Systems},
  volume={131},
  pages={103594},
  year={2020},
  publisher={Elsevier}
}

@article{nava2019direct,
  title={Direct force feedback control and online multi-task optimization for aerial manipulators},
  author={Nava, Gabriele and Sabl{\'e}, Quentin and Tognon, Marco and Pucci, Daniele and Franchi, Antonio},
  journal={IEEE Robotics and Automation Letters},
  volume={5},
  number={2},
  pages={331--338},
  year={2019},
  publisher={IEEE}
}

@article{ruggiero2018aerial,
  title={Aerial manipulation: A literature review},
  author={Ruggiero, Fabio and Lippiello, Vincenzo and Ollero, Anibal},
  journal={IEEE Robotics and Automation Letters},
  volume={3},
  number={3},
  pages={1957--1964},
  year={2018},
  publisher={IEEE}
}

@article{pounds2014stability,
  title={Stability of helicopters in compliant contact under PD-PID control},
  author={Pounds, Paul EI and Dollar, Aaron M},
  journal={IEEE Transactions on Robotics},
  volume={30},
  number={6},
  pages={1472--1486},
  year={2014},
  publisher={IEEE}
}

@inproceedings{ruggiero2015multilayer,
  title={A multilayer control for multirotor UAVs equipped with a servo robot arm},
  author={Ruggiero, Fabio and Trujillo, Miguel Angel and Cano, Raul and Ascorbe, H and Viguria, Antidio and Per{\'e}z, C and Lippiello, Vincenzo and Ollero, An{\'\i}bal and Siciliano, Bruno},
  booktitle={2015 IEEE international conference on robotics and automation (ICRA)},
  pages={4014--4020},
  year={2015},
  organization={IEEE}
}

@article{kamel2018voliro,
  title={Voliro: An omnidirectional hexacopter with tiltable rotors},
  author={Kamel, Mina and Verling, Sebastian and Elkhatib, Omar and Sprecher, Christian and Wulkop, Paula and Taylor, Zachary and Siegwart, Roland and Gilitschenski, Igor},
  journal={arXiv preprint arXiv:1801.04581},
  year={2018}
}

@article{bodie2020active,
  title={Active interaction force control for contact-based inspection with a fully actuated aerial vehicle},
  author={Bodie, Karen and Brunner, Maximilian and Pantic, Michael and Walser, Stefan and Pf{\"a}ndler, Patrick and Angst, Ueli and Siegwart, Roland and Nieto, Juan},
  journal={IEEE Transactions on Robotics},
  volume={37},
  number={3},
  pages={709--722},
  year={2020},
  publisher={IEEE}
}

@article{samadikhoshkho2020nonlinear,
  title={Nonlinear control of aerial manipulation systems},
  author={Samadikhoshkho, Z and Ghorbani, S and Janabi-Sharifi, F and Zareinia, K},
  journal={Aerospace Science and Technology},
  volume={104},
  pages={105945},
  year={2020},
  publisher={Elsevier}
}

@article{lippiello2012cartesian,
  title={Cartesian impedance control of a UAV with a robotic arm},
  author={Lippiello, Vincenzo and Ruggiero, Fabio},
  journal={IFAC Proceedings Volumes},
  volume={45},
  number={22},
  pages={704--709},
  year={2012},
  publisher={Elsevier}
}

@article{kannan2013modeling,
  title={Modeling and control of aerial manipulation vehicle with visual sensor},
  author={Kannan, Somasundar and Olivares-Mendez, Miguel A and Voos, Holger},
  journal={IFAC Proceedings Volumes},
  volume={46},
  number={30},
  pages={303--309},
  year={2013},
  publisher={Elsevier}
}

@incollection{wieber2006holonomy,
  title={Holonomy and nonholonomy in the dynamics of articulated motion},
  author={Wieber, P-B},
  booktitle={Fast Motions in Biomechanics and Robotics: Optimization and Feedback Control},
  pages={411--425},
  year={2006},
  publisher={Springer}
}

@article{wei2021review,
  title={Review of aerial manipulator and its control},
  author={Wei-hong, Xu and Li-jia, Cao and Chun-lai, Zhong},
  journal={International Journal of Robotics and Control Systems},
  volume={1},
  number={3},
  pages={308--325},
  year={2021}
}

@inproceedings{cisneros2018robust,
  title={Robust humanoid control using a QP solver with integral gains},
  author={Cisneros, Rafael and Benallegue, Mehdi and Benallegue, Abdelaziz and Morisawa, Mitsuharu and Audren, Herv{\'e} and Gergondet, Pierre and Escande, Adrien and Kheddar, Abderrahmane and Kanehiro, Fumio},
  booktitle={2018 IEEE/RSJ International Conference on Intelligent Robots and Systems (IROS)},
  pages={7472--7479},
  year={2018},
  organization={IEEE}
}

@article{hua2010attitude,
  title={Attitude estimation for accelerated vehicles using GPS/INS measurements},
  author={Hua, Minh-Duc},
  journal={Control engineering practice},
  volume={18},
  number={7},
  pages={723--732},
  year={2010},
  publisher={Elsevier}
}

@article{rossi2016trajectory,
  title={Trajectory generation for unmanned aerial manipulators through quadratic programming},
  author={Rossi, Roberto and Santamaria-Navarro, Angel and Andrade-Cetto, Juan and Rocco, Paolo},
  journal={IEEE Robotics and Automation Letters},
  volume={2},
  number={2},
  pages={389--396},
  year={2016},
  publisher={IEEE}
}

@article{landau1989applications,
  title={Applications of the passive systems approach to the stability analysis of adaptive controllers for robot manipulators},
  author={Landau, ID and Horowitz, R},
  journal={International Journal of Adaptive Control and Signal Processing},
  volume={3},
  number={1},
  pages={23--38},
  year={1989},
  publisher={Wiley Online Library}
}

@article{garofalo2021adaptive,
  title={Adaptive passivity-based multi-task tracking control for robotic manipulators},
  author={Garofalo, Gianluca and Wu, Xuwei and Ott, Christian},
  journal={IEEE Robotics and Automation Letters},
  volume={6},
  number={4},
  pages={7129--7136},
  year={2021},
  publisher={IEEE}
}

@article{garofalo2020hierarchical,
  title={Hierarchical tracking control with arbitrary task dimensions: Application to trajectory tracking on submanifolds},
  author={Garofalo, Gianluca and Ott, Christian},
  journal={IEEE Robotics and Automation Letters},
  volume={5},
  number={4},
  pages={6153--6160},
  year={2020},
  publisher={IEEE}
}

@inproceedings{lee2010geometric,
  title={Geometric tracking control of a quadrotor UAV on SE (3)},
  author={Lee, Taeyoung and Leok, Melvin and McClamroch, N Harris},
  booktitle={49th IEEE conference on decision and control (CDC)},
  pages={5420--5425},
  year={2010},
  organization={IEEE}
}
\end{document}